\documentclass[sigconf]{acmart}

\AtBeginDocument{%
  }

\usepackage{soul}
\usepackage{xcolor}

\usepackage{amsmath}
\usepackage{amssymb}
\usepackage{mathrsfs}
\usepackage{subcaption}
\usepackage{booktabs}
\usepackage{threeparttable}  % in the preamble
\usepackage{siunitx,booktabs} % (already used above)
\usepackage{bm}

\usepackage[ruled,vlined]{algorithm2e}
\usepackage{enumitem}

\def\model{HAFT}

\setcopyright{rightsretained}
\acmConference[KDD '26]{Proceedings of the 32st ACM SIGKDD Conference on Knowledge Discovery and Data Mining}{August xx--xx, 2026}{Jeju, Korea}
\acmBooktitle{KDD '26: ...}
\acmYear{2026}
\copyrightyear{2026}
\acmDOI{10.1145/....}
\acmISBN{979-8-4007-1454-2/xx/xx}

\begin{document}
\begin{comment}
% Add your initial custom page
\thispagestyle{empty}  % No page number
\begin{center}
    {\Huge\bfseries summary of changes}\\[2cm]
       First, we would like to express our sincere gratitude for the valuable feedback received during the previous round of review, which has helped us further improve the overall framework and quality of our work. A summary of the changes is provided below.\\[1em]

        2. Added the AP-omentum-ovary dataset to test scenarios involving a large number of features.\\[0.5em]
        
        3. Introduced an experiment (in Appendix) to examine whether the framework can trade short-term losses for long-term gains.\\[0.5em]
        
        4. Enhanced the literature review section by incorporating the relevant works suggested by the reviewers.
\end{center}
\clearpage  % Forces content after this to go to the next page
\end{comment}

%remove ACM format information
%\settopmatter{printfolios=false}
%\settopmatter{printacmref=false} % Removes citation information below abstract
%\renewcommand\footnotetextcopyrightpermission[1]{} % removes footnote with conference information in first column
%\pagestyle{empty} % removes running headers

\title{Heterogeneous Multi-Agent Reinforcement Learning with Attention for Cooperative and Scalable Feature Transformation}
\author{Tao Zhe}
\affiliation{%
  \institution{University of Kansas}
  \city{Lawrence}
  \state{KS}
  \country{USA}
}
\email{taozhe@ku.edu}

\author{Huazhen Fang}
\affiliation{%
  \institution{Michigan State University}
  \city{East Lansing}
  \state{MI}
  \country{USA}
}
\email{hfang@msu.edu}

\author{Kunpeng Liu}
\affiliation{%
  \institution{Clemson University}
  \city{Clemson}
  \state{SC}
  \country{USA}
}
\email{kunpenl@clemson.edu}

\author{Qian Lou}
\affiliation{%
  \institution{University of Central Florida}
  \city{Orlando}
  \state{FL}
  \country{USA}
}
\email{qian.Lou@ucf.edu}

\author{Tamzidul Hoque}
\affiliation{%
  \institution{University of Kansas}
  \city{Lawrence}
  \state{KS}
  \country{USA}
}
\email{hoque@ku.edu}

\author{Dongjie Wang}
\authornote{Corresponding author.}
\affiliation{%
  \institution{University of Kansas}
  \city{Lawrence}
  \state{KS}
  \country{USA}
}
\email{wangdongjie@ku.edu}

\begin{abstract}
Feature transformation enhances downstream task performance by generating informative features through mathematical feature crossing. 
% \textcolor{blue}{Qian: As a non-expert in this field, I find the expression "Feature transformation enhances downstream task performance by generating informative features through mathematical feature crossing" more readable.}
Despite the advancements in deep learning, feature transformation remains essential, particularly for structured data, where deep models often struggle to capture complex feature interactions effectively.
% \textcolor{blue}{Qian: I suggest using the expression: "Despite the advancements in deep learning, feature transformation remains essential, particularly for structured data, where deep models often struggle to capture complex feature interactions effectively."}
% \textcolor{blue}{Tao: thank you, professor Qian, indeed much better with your advice."}
Prior literature on automated feature transformation has achieved notable success but often relies on heuristics or exhaustive searches, leading to inefficient and time-consuming processes. 
Recent works employ reinforcement learning (RL) to enhance traditional approaches through a more effective trial-and-error way.
However, two key limitations remain: 
1) Dynamic feature expansion during the transformation process, which introduces instability and increases the time complexity of the learning procedure for RL agents;
2) Insufficient cooperation and communication between agents, which results in suboptimal feature crossing operations and degraded model performance.
To address them, we propose a novel heterogeneous multi-agent RL framework to enable cooperative and scalable feature transformation. 
The framework comprises three heterogeneous agents, grouped into two types, each designed to select essential features and operations for feature crossing. 
To enhance communication among these agents, we implement a shared critic mechanism that facilitates information exchange during the feature transformation process. 
This collaboration enables the agents to learn more intelligent and effective transformation policies.
To handle the dynamically expanding feature space, we tailor multi-head attention-based feature agents to select suitable features for feature crossing.
This design facilitates scalable decision-making and effective candidate selection based on comprehensive global feature space information.
Additionally, we introduce a state encoding technique during the optimization process to stabilize and enhance the learning dynamics of the RL agents, resulting in more robust and reliable transformation policies.
Finally, we conduct extensive experiments to validate the effectiveness, efficiency, robustness, and interpretability of our model. 
Our code and dataset are publicly available on GitHub\footnote{ \url{https://github.com/Ricksanchez000/HAFT}}.
\end{abstract}


%\settopmatter{printacmref=false}
% \setcopyright{none}
% \renewcommand\footnotetextcopyrightpermission[1]{} % Remove footnote with conference info
% \acmConference{}{}{}  % Blank out conference data
% \acmYear{}
% \copyrightyear{}
% \acmDOI{}
% \acmISBN{}

% 关键词

% CCS 分类（从 https://dl.acm.org/ccs 生成）


\maketitle

%%%%%%%%%%%%%%%%%%%%%%%%%%
\section{Introduction}
Feature transformation creates more informative features by mathematically transforming the original ones. 
This enhances pattern distinguishability and improves model performance.
Despite the remarkable success of deep learning, feature transformation remains essential for structured data (e.g., tabular data)~\cite{fu2024tabular} and scenarios requiring high interpretability or the integration of domain-specific knowledge.
For instance, in finance, features like debt-to-income ratio and credit utilization are derived from raw data. These features enhance interpretability and improve the performance of tasks such as credit risk assessment and loan approval.
In healthcare, risk scores are derived by transforming clinical measurements into interpretable features. These features support more accurate diagnosis and treatment planning.
All of these examples demonstrate that feature transformation still plays an important role in real-world scenarios and AI applications.

Existing work on automated feature transformation can be categorized:
1) Expansion reduction methods~\cite{kanter2015deep,horn2020autofeat,khurana2016cognito}, which expand the feature space by applying mathematical operations and then select the most suitable features to reduce the feature space size; 
2) Iterative feedback loops~\cite{wang2022group, khurana2018feature, tran2016genetic,huang2025collaborative,hu2024reinforcement}, which iteratively refine features based on feedback from downstream tasks, optimized by RL or evolutionary algorithms;
3) AutoML-based methods~\cite{chen2019neural, zhu2022difer,gao2025gpt,liu2025continuous,wang2023reinforcement,xiao2023beyond}, which frame feature transformation as a search problem for neural architectures and solve it using AutoML techniques.
Among these, iterative-feedback paradigms excel at dynamically refining features over iterations.
They can adapt to diverse downstream tasks and often yield effective feature spaces.

Recent studies have shown the effectiveness of RL in feature transformation~\cite{wang2022group,hu2024reinforcement,huang2024enhancing,ying2023self,xiao2024traceable,ying2025survey,xiao2023traceable}.
Typically, RL agents are employed to select the necessary features and mathematical operations to generate new informative features. 
This process is repeated iteratively with the goal of optimizing downstream task performance, continuing until the maximum iteration limit is reached.
However, existing RL-based methods still face two key challenges:
\begin{itemize}

 \item \textbf{Challenge 1: Continuous dynamic expansion of the feature space during the iterative RL learning process.}
    During the feature transformation, the feature space undergoes continuous dynamic expansion over time. This expansion makes RL agents difficult and inefficient in identifying key features that are required for effective feature crossing. 
    
    \item \textbf{Challenge 2:  Insufficient cooperation and communication among different RL agents for policy learning.}
    Existing works often rely on purely local exchanges (e.g., selected features or operations from previous agents), limiting the agents' awareness of the global feature space and resulting in suboptimal policies and feature spaces.
    
\end{itemize}

\begin{figure*}[!thbp]
  \centering
  \includegraphics[width=\linewidth]{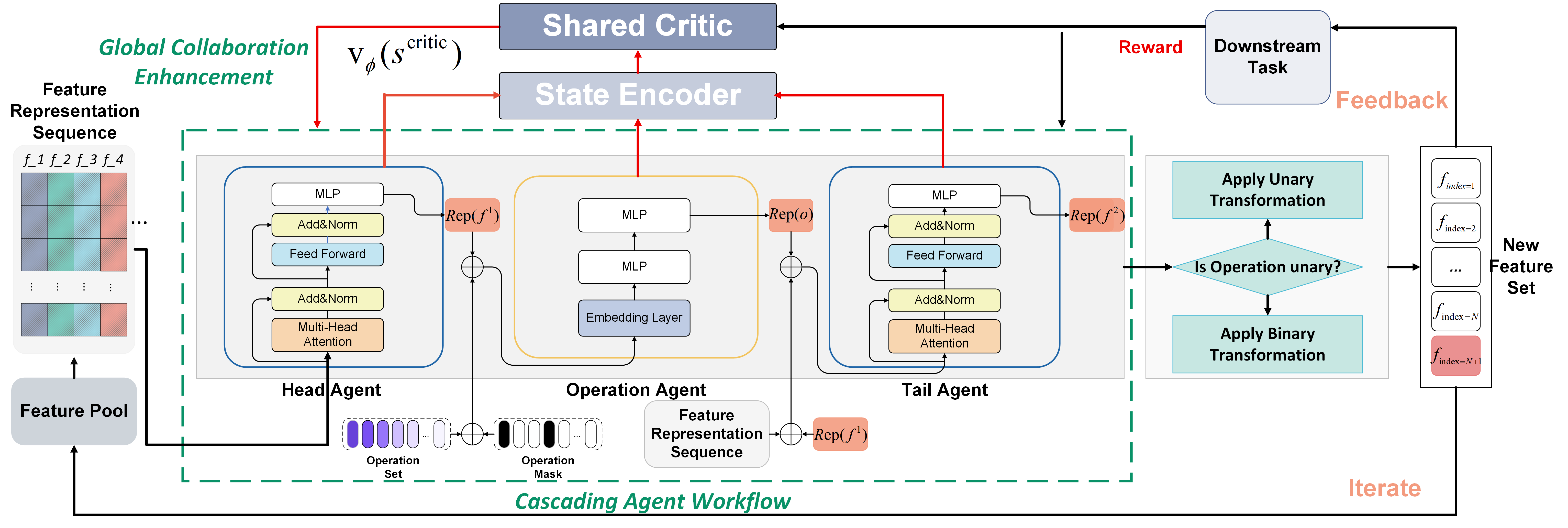}
  \caption{Framework Overview. 
  \model\ consists of three heterogeneous agents: two feature agents that select candidate features and one operation agent that determines a mathematical operation.
  A central critic leverages global information from the current feature space to coordinate the agents and enhance policy learning. This feature transformation process continues iteratively until an optimal feature set is identified or the iteration limit is reached.}
  \Description{An overview of our framework}
  \label{Framework Overview}
  \vspace{-0.4cm}
\end{figure*}

To address these challenges, we propose a \textbf{\underline{H}}eterogeneous multi-\textbf{\underline{A}}gent reinforcement learning framework for cooperative and scalable \textbf{\underline{F}}eature \textbf{\underline{T}}ransformation, referred to as \textbf{\model}.
Specifically, we first design three RL agents, with two responsible for choosing candidate features and one for choosing mathematical operations. 
With feature choice facing an expanding space and operation choice using a static set, we develop two types of heterogeneous agents.
This design ensures that each task is handled efficiently and effectively, enhancing the scalability and adaptability of the feature transformation process.
Then, we propose an attention-based feature agent structure to adapt to the dynamically expanding feature space and identify the most suitable feature for the next feature crossing step. By focusing on relevant features and incorporating global context, this structure allows the agent to capture dynamic changes in the feature space and understand complex feature interactions for decision making.
Moreover, we introduce a shared central critic for the RL agents to evaluate their decisions using unified information throughout the transformation process. 
By providing a global view of the feature space and incorporating the actions of other agents, the critic enhances communication and cooperation among them. 
This strategy enables each agent to make more informed and coordinated decisions, fostering effective feature transformation policies. 
To further stabilize and enhance the learning process, we propose a state encoding technique. 
This method mitigates significant state changes caused by feature space expansion, which stabilizes RL learning and improves the feature space exploration process.

To summarize, our contributions are threefold:
\begin{itemize}
    \item \textbf{\textit{Problem:}}  We propose a new framework for automated feature transformation that leverages multi-agent cooperation to efficiently and scalably explore the feature space.
    \item \textbf{\textit{Algorithm:}} We design a heterogeneous agent structure, customized based on the task of selecting features or operations. 
    An attention-based feature agent structure is implemented to adapt to the dynamically expanding feature space. Moreover, we propose a shared critic structure to enhance communication and cooperation among the agents.
    \item \textbf{\textit{Evaluation:}} We conduct extensive experiments on 23 real-world datasets to validate the effectiveness of \model. 
    The experimental results demonstrate the superiority of \model\ and highlight the indispensability of each of its components.
\end{itemize}

\section{Problem Statement}
We aim to develop an automated feature transformation framework that efficiently refines the feature space to maximize downstream task performance.
Given a dataset $\mathcal{D}\langle \mathcal{F}, y\rangle$, where $\mathcal{F}$ represents input features and $y$ is the target variable, and a mathematical operation set $\mathcal{O}$ (e.g., addition, multiplication, division), the objective is to generate an optimal feature set $\mathcal{F}^*$ that maximizes ML task $A$'s performance metric $P^A$.
The optimization goal can be formulated as follows: 
\begin{equation}
\mathcal{F}^* = \arg\max_{\hat{\mathcal{F}}} \left( P^A \big( \hat{\mathcal{F}}, y \big) \right),
\end{equation}
where $\hat{\mathcal{F}}$ is a feature subset derived from the union of the transformed new features $\mathcal{F}^g$ and the original feature set $\mathcal{F}$.
$\mathcal{F}^g$ is generated by applying the mathematical operations from $\mathcal{O}$ to features from the original feature set $\mathcal{F}$ for feature crossing.
For example, multiplying \textit{length} and \textit{width} creates a new feature \textit{area=length$\times$width}. 
After reaching the maximum iteration limit, the feature space with the highest downstream task performance is output as $\mathcal{F}^*$.

%%%%%%%%%%%%%%%%%%%%%

%\vspace{-0.2cm}
\section{Methodology}
\label{sec:proposed_method}
% In this section, we present an overview, and then detail each technical component of our framework.

\subsection{Framework Overview}
\label{sec:framework_overview}

Figure~\ref{Framework Overview} demonstrates the architecture of \model. In the iterative optimization process, three heterogeneous cascading agents collaborate to learn the feature transformation policy. 
Two feature agents identify candidate features, while the operation agent selects a mathematical operation. 
Given the continuous evolution of the feature space, \model\ incorporates a multi-head attention-based feature agent structure to select features that contribute effectively to feature transformation using global information.
This design maintains a consistent agent structure, ensuring effectiveness and scalability without structural modifications.
To manage the complex dependencies among agents, a shared central critic integrates global feature space information and agent decisions to guide and enhance policy learning. 
The selected operation is then applied to the candidate features to generate new and informative features in each iteration.
The optimization process will be guided by the performance of downstream tasks. 
The entire process is repeated until either the optimal feature space is identified or the maximum iteration limit is reached.
Finally, the feature space with the highest downstream task performance is output as the optimal solution.

%%%%%%%%%%%%%%%%%%%
\subsection{Heterogeneous Cascading Agents for Candidate Features and Operations}

To handle dynamic and complex nature of automated feature transformation, we design a cascading structure with three heterogeneous agents: a head feature agent, an operation agent, and a tail feature agent. 
They collaborate iteratively to identify candidate features and operations for feature transformation.
Each agent is specialized for a specific role to better capture global context, grasp dependencies, and adapt to the dynamic evolving feature spaces.

\textbf{Cascading Agent Workflow:}
At the $t$-th iteration, the three agents collaborate to choose candidate features and operation for feature crossing.
In the following, we first outline the workflow and define the state, action, and reward for each agent in the process.

\noindent\ul{Head feature agent} selects a feature $f^1_{t-1}$ from the feature set $\mathcal{F}_{t-1}$ based on the information of global feature space.
Within this agent, \textbf{state:} $s_t^1=\text{Rep}(\mathcal{F}_{t-1})$, where $\text{Rep}(\mathcal{F}_{t-1})$ is the representation of the current feature space.
Here, $\text{Rep}(\mathcal{F}_{t-1})$ is obtained by statistical descriptors of the feature set (e.g., mean and variance).
\textbf{action:} $a_t^1=f^1_{t-1}$ is the selected feature from the feature set $\mathcal{F}_{t-1}$.

\noindent\ul{Operation agent} selects an operation $o_{t}$ from the operation set $\mathcal{O}$ based on the decision of the head feature agent and the current dynamic mask(in equation~\ref{eq: dynamic_mask}).
Within this agent, \textbf{state:} $s_t^o=\mathbf{c}_{t} \oplus \text{Rep}(f^1_{t-1})$, which is the concatenation of the representation of the current dynamic mask and the first candidate feature.
\textbf{action:} $a_t^o=o_{t}$ is the selected feature from the mathematical operation set $\mathcal{O}$.

\noindent\ul{Tail feature agent} selects second feature $f^2_{t-1}$ from the feature set $\mathcal{F}_{t-1}$  based on the decisions of the head feature agent, operation agent, and the current feature space representation.
Within this agent, \textbf{state:} $s_t^2=\text{Rep}(\mathcal{F}_{t-1}) \oplus \text{Rep}(f^1_{t-1}) \oplus \text{Rep}(o_{t}) $, where $\text{Rep}(o_{t})$ is the one-hot encoding vector for the selected operation.
\textbf{action:} $a_t^2=f^2_{t-1}$ is the selected feature from the feature set $\mathcal{F}_{t-1}$.

\noindent\ul{Reward:} The three agents are all driven by the downstream machine learning task performance $V_A$. They equally share the reward, defined as $\mathcal{R}(s_t^1,a_t^1) = \mathcal{R}(s_t^o,a_t^o)= \mathcal{R}(s_t^2,a_t^2) = P^A_t$.

\subsubsection{Multi-Head Attention-Based Feature Agent}
%\label{sec:transformer_encoder}

\noindent \textbf{Why dynamic evolving feature space matters?} 
During the iterative process, the size of the feature space changes after each iteration.
This dynamic nature makes it challenging to design an agent structure that can handle variable input sizes and capture complex feature interactions effectively.
Prior research~\cite{wang2022group} employs group-wise clustering to address the dynamic changes within the feature space. 
While effective in capturing high-level group structures, this approach struggles with scalability and modeling fine-grained feature relationships.
Thus, we propose a multi-head attention-based~\cite{vaswani2017attention} agent structure to address these limitations.
In the following, we show the calculation process using the feature space at iteration $t$. To simplify notation, we omit $t$, using $\mathcal{F}$ to represent $\mathcal{F}_t$.

Let $\mathbf{H}_0 = \textit{Rep}(\mathcal{F}) \in \mathbb{R}^{N \times d}$ be the initial feature matrix, where $N$ is the number of features and $d$ is the representation dimension.
A Transformer Encoder Layer applies multi-head self-attention to the feature matrix to capture the complex interaction between features. Concretely, use $\mathbf{H}_k$ to denote the hidden representations after the $k^\text{th}$ encoder block. Each block has $h$ attention heads. For head $j \in \{1,2,\dots,h\}$, we compute:
\(\mathbf{Q}_k^j \;=\; \mathbf{H}_{k-1} \,\mathbf{W}_{k,Q}^j\), \(\mathbf{K}_k^j \;=\; \mathbf{H}_{k-1} \,\mathbf{W}_{k,K}^j\), and \(\mathbf{V}_k^j \;=\; \mathbf{H}_{k-1} \,\mathbf{W}_{k,V}^j\),
where $\mathbf{W}_{k,Q}^j, \mathbf{W}_{k,K}^j, \mathbf{W}_{k,V}^j \in \mathbb{R}^{d \times (d/h)}$ are learnable projection matrices for queries, keys, and values. By using different learned projection parameters to generate distinct sets of queries, keys, and values, Each head operates independently on these projections, enabling the model to capture various aspects of the relationships among features. The scaled dot-product attention for head $j$ is given by:
    \(\mathbf{H}_k^j \;=\; \text{softmax}\!\Bigl(
    \frac{\mathbf{Q}_k^j \,(\mathbf{K}_k^j)^\top}{\sqrt{d/h}}
    \Bigr)\;\mathbf{V}_k^j.\)
%\paragraph{Feed-Forward Update.}
We then concatenate the outputs and pass them to a feed-forward function:

\begin{equation}
    \mathbf{H}_{k}
    \;=\;
    \mathrm{FFN}\!\left(
        \operatorname{Concat}\bigl(
            \mathbf{H}_{k}^{1}, \dots, \mathbf{H}_{k}^{h}
        \bigr)
    \right).
\end{equation}

This yields the representation for the next layer. Stacking $K$ such blocks results in the final encoder output
\(\mathbf{H}_K\).
We input $\mathbf{H}_K$ into a linear layer with softmax activation function to learn a distribution over all features.
The feature agent samples the optimal feature for crossing based on the learned distribution.
This design adapts to dynamic changes in the feature space, as self-attention assigns weights based on feature relevance without requiring a fixed input size or manual indexing, ensuring scalability for large feature space.

\subsubsection{MLP-Based Operation Agent}
\label{sec:operation_agent}
Unlike feature agents, the operation agent selects an operation from a fixed set, making its task simpler than choosing features. 
To accommodate this difference, we design an MLP-based operation agent. 
This heterogeneity in agent structure allows task-specific customization, which enhances policy learning and improves both feature and operation selection efficiency.
Since the input size is fixed, we design an MLP-based operation policy network with a \textit{dynamic mask} to enable both informed and valid operation selection. 
Certain operations may be invalid for specific features (e.g., \texttt{sqrt} is invalid if the feature contains negative values). To handle this, we define a dynamic action mask $\mathbf{c} \in \{0,1\}^m$, where $m$ is the number of predefined operations:
\begin{equation}
\label{eq: dynamic_mask}
\mathbf{c} = (c_1, c_2, \dots, c_m), \quad
c_i =
\begin{cases}
    1, & \text{if the $i$-th operation is valid},\\
    0, & \text{otherwise}.
\end{cases}
\end{equation}
The input of the policy network is the concatenation of the selected head feature representation and the mask \(\mathbf{c}\). This combined input is processed by a feed-forward neural network that outputs logits for each operation. 
For any invalid operation (i.e., \(c_i = 0\)), a large negative constant is added to its logit, effectively reducing its selection probability to near zero. This mechanism prevents invalid operations and overly deterministic policies, ensuring valid transformations and a more stable policy-learning process.

%\subsection{Shared Critic for Collaborative Feature Transformation}
\subsection{Enhancing Collaboration with a Shared Critic and Advantage Decomposition}
\label{sec:central_critic_sec}
\textbf{Why a shared central critic matters?} 
Existing works~\cite{wang2022group} on automated feature transformation rely on partial information (e.g., decisions from previous agents) to coordinate and learn transformation policies.
However, convergence in this setting is particularly challenging, as each agent's policy is influenced not only by preceding agents but also by subsequent ones, creating complex interdependencies.
To improve communication among agents and learn better feature transformation policies, a shared central critic module is essential. 
This necessitates robust credit assignment and alignment mechanisms.
Unlike conventional multi-agent systems or feature-selection RL—where action spaces remain fixed and counterfactual baselines are straightforward to define—feature transformation presents unique difficulties. The dynamic expansion of the feature pool throughout the learning process makes credit assignment substantially more complex, as each agent's contribution becomes increasingly entangled with the evolving feature space. This contrasts with typical multi-agent settings where individual contributions to shared rewards are more readily identifiable or where counterfactual reasoning can be more directly applied.

To address these unique challenges, we apply a \textbf{Shared Critic} with \textbf{multi-agent advantage decomposition and a sequential policy-update scheme}\cite{kuba2021trust,yu2022surprising}. The critic consumes aggregated global information and outputs a single scalar value estimate:
\begin{equation}
\hat{V}_t = V_{\phi}\left(s^{\text{critic}}\right),
\end{equation}
which serves as a unified baseline for advantage estimation. Each agent (head feature agent, tail feature agent, and operation agent) still acts on its local observation, but the centralized perspective of the Shared Critic aligns their learning toward producing high-quality transformations. With advantage decomposition, subtracting $\hat{V}_t$ from individual returns yields clearer reward attribution, stabilizes training, reduces non‑stationarity, and strengthens collaboration. Building on this design, we now detail each agent’s update procedure within our framework.

\begin{comment}
\begin{multline}
\mathbb{E}_{s \sim \rho_{\pi_{\theta_k}}, a^i \sim \pi_{\theta_k}} \Bigg[ \min \Bigg( \frac{\pi_{\theta_{im}}(a^i|s)}{\pi_{\theta_k^{im}}(a^i|s)} M^{i1:m}(s,a), \\
\text{clip}\left(\frac{\pi_{\theta_{im}}(a^i|s)}{\pi_{\theta_k^{im}}(a^i|s)}, 1 \pm \varepsilon \right) M^{i1:m}(s,a) \Bigg) \Bigg].
\end{multline}
\end{comment}

\noindent \textbf{Agents Update}.
We adopt the customized HAPPO algorithm to update our three agents, as it effectively addresses the complexity and instability inherent in our dynamic multi-agent environment while maintaining theoretical monotonic improvement guarantees. Its clipped objective function confines policy updates within a trust region, preventing destabilizing shifts during training. 
We employ the Generalized Advantage Estimation (GAE) advantage function to compute the joint advantage function \(\hat{A} \), which balances bias and variance through its exponential weighting scheme.
\begin{comment}
First, we compute the temporal difference (TD) error at each time step:
\(
\delta_t = r_t + \gamma \hat{V}_{\phi}(s_{t+1}) - \hat{V}_{\phi}(s_t)
\)
where $r_t$ is the immediate reward at time $t$, $\gamma$ is the discount factor, and $\hat{V}_{\phi}(s_t)$ represents the centralized value network with parameters $\phi$.
The GAE advantage estimator is then computed as an exponentially-weighted average of TD errors:
where $\lambda \in [0,1]$ is the GAE parameter that controls the bias-variance tradeoff.
\end{comment}
To utilize the advantage decomposition in our setting, we sequentially update our agents. While the order can be randomized in theory, we fix a consistent order (head feature agent $\rightarrow$ operation agent $\rightarrow$ tail feature agent) for simplicity.
For the first agent in the sequence (e.g., the feature agent), we define:
\(
M_t^{\mathrm{^1}} = \hat{A}_t,
\)
and the importance-sampling ratio:
\(
\rho_t^{\mathrm{^1}}(\theta^{\mathrm{^1}}) = 
\frac{\pi_{\theta^{\mathrm{^1}}}(a_t^{\mathrm{^1}} \mid s_t^{\mathrm{^1}} )}
     {\pi_{\theta^{\mathrm{^1}}_{\text{old}}}(a_t^{\mathrm{^1}} \mid s_t^{\mathrm{^1}} )}.
\)
The head feature agent’s clipped objective is then:
\begin{multline}
L_{\mathrm{^1}}(\theta^{\mathrm{^1}})
= \mathbb{E}_t\Bigl[
   \min\bigl(
     \rho_t^{\mathrm{^1}}(\theta^{\mathrm{^1}})\,M_t^{\mathrm{^1}},
     \mathrm{clip}\bigl(\rho_t^{\mathrm{^1}}(\theta^{\mathrm{^1}}),\,1-\epsilon,\,1+\epsilon\bigr)\,M_t^{\mathrm{^1}}
   \bigr)
\Bigr].
\end{multline}
After updating the feature agent’s parameters $\theta^{\mathrm{^1}}_{\text{old}} \rightarrow \theta^{\mathrm{^1}}_{\text{new}}$, we adjust the advantage for the next agent (e.g., the operation agent) to account for the updated policy:
\(
M_t^{\text{op}} = 
\frac{\pi_{\theta^{\mathrm{^1}}_{\text{new}}}(a_t^{\mathrm{^1}} \mid s_t^{\mathrm{^1}} )}
     {\pi_{\theta^{\mathrm{^1}}_{\text{old}}}(a_t^{\mathrm{^1}} \mid s_t^{\mathrm{^1}} )} \cdot \hat{A}_t.
\)
The operation agent's ratio and objective are defined analogously, following the same clipped formulation.
\( \rho_t^{\mathrm{o}}(\theta^{\mathrm{o}})
    = \frac{\pi_{\theta^{\mathrm{o}}}(a_t^{\mathrm{o}}\mid s_t^{\mathrm{o}} )}
           {\pi_{\theta_{\mathrm{old}}^{\mathrm{o}}}(a_t^{\mathrm{o}}\mid s_t^{\mathrm{o}})} \),
\begin{multline}
L_{\mathrm{o}}(\theta^{\mathrm{o}})
= \mathbb{E}_t\Bigl[
   \min\bigl(
     \rho_t^{\mathrm{o}}(\theta^{\mathrm{o}})\,M_t^{\mathrm{o}},
     \mathrm{clip}\bigl(\rho_t^{\mathrm{o}}(\theta^{\mathrm{o}}),\,1-\epsilon,\,1+\epsilon\bigr)\,M_t^{\mathrm{o}}
   \bigr)
\Bigr]
\end{multline}
After updating the operation agent’s parameters 
$\theta^{\text{o}}_{\text{old}} \rightarrow \theta^{\text{o}}_{\text{new}}$, 
we form the tail feature agent’s decomposed advantage:
\(
M_t^{\mathrm{^2}} =
\frac{\pi_{\theta^{\mathrm{^1}}_{\text{new}}}(a_t^{\mathrm{^1}} \mid s_t^{\mathrm{^1}})}
     {\pi_{\theta^{\mathrm{^1}}_{\text{old}}}(a_t^{\mathrm{^1}} \mid s_t^{\mathrm{^1}})}
\cdot
\frac{\pi_{\theta^{\mathrm{o}}_{\text{new}}}(a_t^{\mathrm{o}} \mid s_t^{\mathrm{o}})}
     {\pi_{\theta^{\mathrm{o}}_{\text{old}}}(a_t^{\mathrm{o}} \mid s_t^{\mathrm{o}})}
\cdot \hat{A}_t.
\)
Define its ratio:
\(
\rho_t^{\mathrm{^2}}(\theta^{\mathrm{^2}}) =
\frac{\pi_{\theta^{\mathrm{^2}}_{\text{}}}(a_t^{\mathrm{^2}} \mid s_t^{\mathrm{^2}})}
     {\pi_{\theta^{\mathrm{^2}}_{\text{old}}}(a_t^{\mathrm{^2}} \mid s_t^{\mathrm{^2}})}.
\)
The tail feature agent’s clipped objective becomes:
\begin{multline}
L_{\mathrm{^2}}(\theta^{\mathrm{^2}}) =
\mathbb{E}_t \Bigl[
\min\bigl(
  \rho_t^{\mathrm{^2}}(\theta^{\mathrm{^2}})\, M_t^{\mathrm{^2}},
  \mathrm{clip}\bigl(\rho_t^{\mathrm{^2}}(\theta^{\mathrm{^2}}),\,1-\epsilon,\,1+\epsilon\bigr)\, M_t^{\mathrm{^2}}
\bigr)
\Bigr].
\end{multline}

This sequential scheme ensures that each agent’s update accounts for the changes introduced by previously updated agents, mitigating the challenge of conflicting policy update of each agent.
To refine the reward for our feature transformation task, we incorporate characteristics of the generated feature space. Specifically, an ideal feature space generally has low information redundancy and high relevance to the predictive labels. 
We quantify both using mutual information $MI$:
The \textbf{Information redundancy} is defined as the average pairwise $MI$ among features:
\(
    Id = \frac{1}{|N|^2} \sum_{f_i, f_j \in N} MI(f_i; f_j),
\)
where \( N \) is the feature subset, \( f_i \) is the \( i \)-th feature.
The \textbf{information relevance} is the average $MI$ between features and labels:
\(
    Iv = \frac{1}{|N|} \sum_{f_i \in N} MI(f_i; y),
\)
where \( y \) is the label vector.
To encourage exploration, we also add an entropy term $H(\pi_{\theta^{(i)}})$. Our final loss function for optimizing the three agents is thus defined as follows:
For each agent $i \in \{\text{feat\_h}, \text{op}, \text{feat\_t}\}$, the policy loss is given by:
\begin{equation}
\mathcal{L}_{\text{policy}}^{(i)}(\theta^{(i)}) 
= - L_{\text{CLIP}}^{(i)} 
  - \beta_i \, \mathbb{E}_t \big[ H(\pi_{\theta^{(i)}}) \big]
  + \beta_{\text{Id}} \, Id 
  - \beta_{\text{Iv}} \, Iv,
\end{equation}
where $L_{\text{CLIP}}^{(i)}$ is one of the clipped surrogates defined above, $\beta_i$ controls entropy regularization.
And finally the agent's policy network parameters are updated via gradient descent optimization.

\noindent \textbf{Shared Critic Update}. The shared critic is updated by minimizing the mean-squared error (MSE) between the predicted value and the computed returns from trajectories:
\begin{equation}
    \mathcal{L}_{\text{value}}(\phi) = \frac{1}{2}\, \mathbb{E}_t \left[ \left(V_\phi(s_t^{critic}) - \hat{R}_t\right)^2 \right],
\end{equation}
where $\hat{R}_t$ represents the returns computed using GAE. Specifically, $\hat{R}_t = \hat{A}_t + V_\phi(s_t^{critic})$, where $\hat{A}_t$ is the GAE advantage estimate computed from the collected trajectory rewards and bootstrapped values. The network parameters $\phi$ are updated via gradient descent.

\begin{figure}[t]
  \centering
  \includegraphics[width=\linewidth]{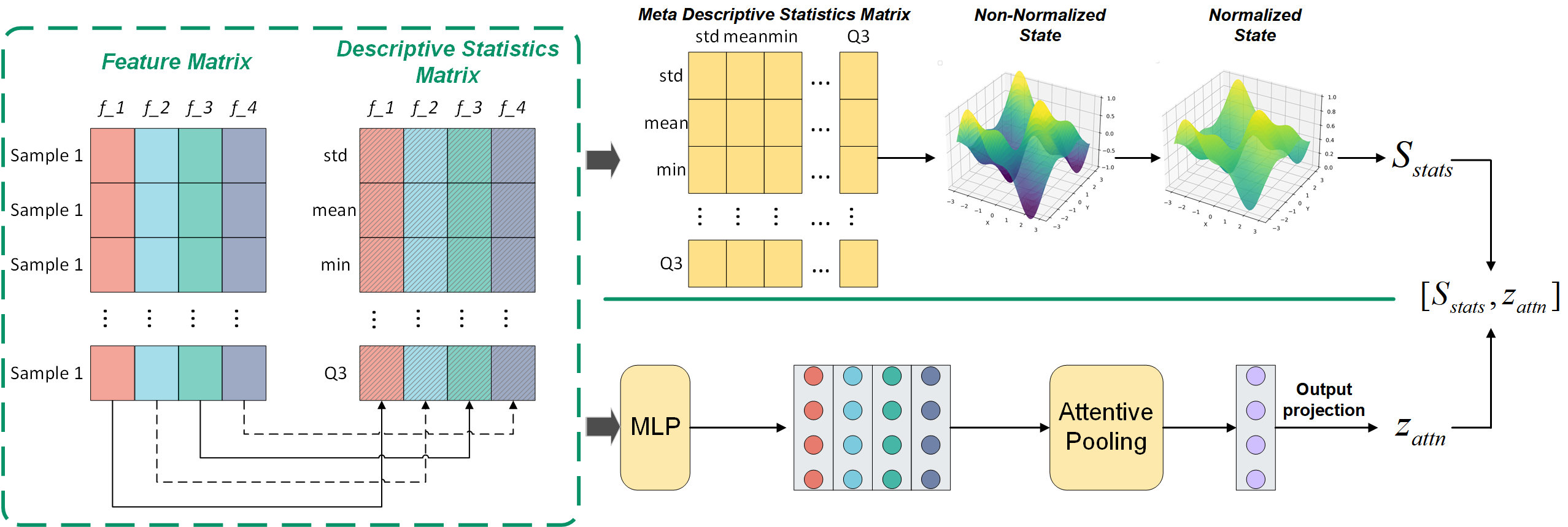}
  \caption{Two‑branch state encoder. A statistics branch expands the dynamic feature set and normalizes it to $\mathbb{R}^{1 \times 49}$; an attention branch embeds features, applies multi‑head attention, and pooling to obtain
  \( z_{attn}\). Finally create input \(s^{\text{critic}} = [S_{\text{stats}}, z_{\text{attn}}]\).}
  \Description{Central Critic Input State.}
  \label{fig:central_critic_input}
  \vspace{-0.4cm}
\end{figure}

%\subsection{Normalized Shared Critic State}
\subsection{State Encoding for Shared Critic} 
\label{normal_state}
As outlined in the shared critic section, the aggregated global information of our multi-agent feature transformation system represents the current feature set environment after each iteration, reflecting the updated features generated by the agents.
Specifically, as the number of features
increases by one after each step, the resulting shift in the input state distribution can destabilize the critic’s evaluation process. This
instability may manifest as erratic output values and, in extreme cases, exploding gradients, which can corrupt the overall training
of the framework.
In addition to these aggregates, heuristic signals are also essential for reliable evaluation.
To provide the critic with a stable yet expressive view of the evolving space, we construct a two‑branch, permutation‑robust state
as shown in Fig.~\ref{fig:central_critic_input}:
i) a distributional‑statistics branch that converts the variable‑length pool into a fixed, normalized summary \( S_{stats} \in \mathbb{R}^{1 \times 49} \), 
ii) an attention‑based interaction branch that embeds per‑feature tokens with multi‑head attention and permutation‑invariant pooling to produce \( z_{attn} \in \mathbb{R}^{1 \times 49} \).
The final critic input is the concatenation :
\begin{equation}
s^{\text{critic}} = [S_{\text{stats}}, z_{\text{attn}}]
\end{equation}
\noindent \textbf{Distributional Statistics \& Normalization}. 
To construct \( S_{stats} \), we propose a two-stage approach.
First, we expand the dynamic feature set by computing a series of descriptive statistics, including the mean, standard deviation, minimum, maximum, and the first, second and third quartiles. Following the first step, we calculate the descriptive statistics again row by row. This statistical summary effectively condenses the raw, variable-length feature set into a fixed set of informative metrics. Second, we apply a normalization step (e.g., min-max normalization or z-score normalization) to these statistics. This normalization is crucial as it scales each statistic to a similar range, preventing any single metric from disproportionately affecting the critic’s output and helping to control gradient magnitudes.
The combined process produces a fixed input state representation with a shape of \( \mathbb{R}^{1 \times 49} \).

\noindent \textbf{Attention‑based Interaction}.
While aggregated global statistics summarize each feature, they do not capture feature–feature interactions, which are critical for constructing a refined feature space. Therefore, we construct  \(z_{attn}\) by adopting an attention‑based encoder: each feature is first mapped to a continuous token embedding; the set of embeddings is then processed with multi‑head attention to produce context‑aware representations. A permutation‑invariant pooling operation (e.g., attention pooling or mean–max pooling) aggregates them into a fixed‑length vector, which is concatenated with the statistics‑based summary to form the shared critic’s state.

The resulting state representation stabilizes training by fixing the input dimensionality regardless of dynamic feature size, and normalizing scales so no statistic dominates the gradients. In practice, it provides a compact, informative signal that improves value estimation and credit assignment, simplifies the critic architecture, and leads to more reliable convergence for our system.

\begin{table*}[!htbp]
\caption{Overall Performance. `C' for classification and `R' for regression.}
\vspace{-0.3cm}
  \label{tab:overall_performance}
  \begin{tabular}{ccccccccccccccc}
    \toprule
    Dataset & Source & C/R & Samples & Features & RDG & ERG & LDA & AFAT & NFS & TTG & GRFG & DIFER & HAFT \\
    \midrule
    Higgs Boson& UCIrvine& C& 50000& 28& 0.695& 0.702& 0.513& 0.697& 0.691& 0.699& \underline{0.707}& 0.669& \textbf{0.709} \\
    Amazon Employee& Kaggle& C& 32769& 9& 0.932& 0.932& 0.916& 0.930& 0.932& \underline{0.933}& 0.932& 0.929& \textbf{0.943} \\
    PimaIndian& UCIrvine& C& 768& 8& 0.751& 0.735& 0.729& 0.732& \underline{0.791}& 0.745& 0.754& 0.760& \textbf{0.794} \\
    SpectF& UCIrvine& C& 267& 44& 0.760& 0.788& 0.665& 0.760& \underline{0.792}& 0.760& 0.776& 0.766& \textbf{0.876} \\
    SVMGuide3& LibSVM& C& 1243& 21& 0.806& \underline{0.818}& 0.635& 0.794& 0.786& 0.798& 0.812& 0.773& \textbf{0.837} \\
    German Credit& UCIrvine& C& 1001& 24& \underline{0.701}& 0.662& 0.596& 0.639& 0.649& 0.644& 0.667& 0.656& \textbf{0.740} \\
    Credit Default& UCIrvine& C& 30000& 25& 0.802& 0.803& 0.743& 0.804& 0.801& 0.798& \underline{0.806}& 0.796& \textbf{0.807} \\
    Messidor\_features& UCIrvine& C& 1150& 19& 0.627& \underline{0.692}& 0.463& 0.657& 0.650& 0.655& 0.680& 0.660& \textbf{0.730} \\
    Wine Quality Red& UCIrvine& C& 999& 12& \underline{0.496}& 0.485& 0.401& 0.480& 0.451& 0.467& 0.454& 0.476&  \textbf{0.561} \\
    Wine Quality White& UCIrvine& C& 4900& 12& 0.524& \underline{0.527}& 0.438& 0.516& 0.525& 0.521& 0.518& 0.507& \textbf{0.543} \\
    SpamBase& UCIrvine& C& 4601& 57& 0.906& 0.917& 0.889& 0.912& \underline{0.925}& 0.919& 0.922& 0.912& \textbf{0.928} \\
    AP-omentum-ovary& OpenML& C& 275& 10936& 0.820& \underline{0.849}& 0.716& 0.813& 0.830& 0.830& 0.830& 0.833& \textbf{0.879} \\
    Lymphography& UCIrvine& C& 148& 18& 0.108& \underline{0.202}& 0.144& 0.149& 0.166& 0.148& 0.136& 0.150& \textbf{0.397} \\
    Ionosphere& UCIrvine& C& 351& 34& 0.942& \underline{0.957}& 0.743& 0.942& 0.943& 0.932& 0.956& 0.905& \textbf{0.957} \\
    Housing Boston& UCIrvine& R& 506& 13& \underline{0.434}& 0.410& 0.020& 0.401& 0.428& 0.396& 0.409& 0.381& \textbf{0.476} \\
    Airfoil& UCIrvine& R& 1503& 5& 0.519& 0.519& 0.207& 0.507& 0.520& 0.500& 0.521& \underline{0.528}& \textbf{0.548} \\
    Openml\_618& OpenML& R& 1000& 50& 0.472& \underline{0.561}& 0.052& 0.473& 0.467& 0.467& \textbf{0.577}& 0.408& 0.475 \\
    Openml\_589& OpenML& R& 1000& 25& 0.509& 0.610& 0.011& 0.508& 0.503& 0.504& \textbf{0.627}& 0.463& \underline{0.555} \\
    Openml\_616& OpenML& R& 500& 50& 0.070& 0.193& 0.024& 0.162& 0.147& 0.156& \textbf{0.372}& 0.076& \underline{0.271} \\
    Openml\_607& OpenML& R& 1000& 50& 0.521& 0.549& 0.107& 0.509& 0.518& 0.522& \underline{0.555}& 0.476& \textbf{0.618} \\
    Openml\_620& OpenML& R& 1000& 25& 0.511& 0.546& 0.029& 0.527& 0.513& 0.512& \textbf{0.619}& 0.442& \underline{0.618} \\
    Openml\_637& OpenML& R& 500& 50& 0.136& 0.064& 0.043& 0.161& 0.146& 0.144& \textbf{0.307}& 0.072& \underline{0.300} \\
    Openml\_586& OpenML& R& 1000& 25& 0.568& 0.624& 0.110& 0.553& 0.542& 0.544& \underline{0.646}& 0.482& \textbf{0.707} \\
    
    \bottomrule
  \end{tabular}
\end{table*}
%\vspace{-0.7cm}

\begin{figure*}[h]
  \centering
    \begin{subfigure}[t]{0.24\textwidth}
        \includegraphics[width=\textwidth]{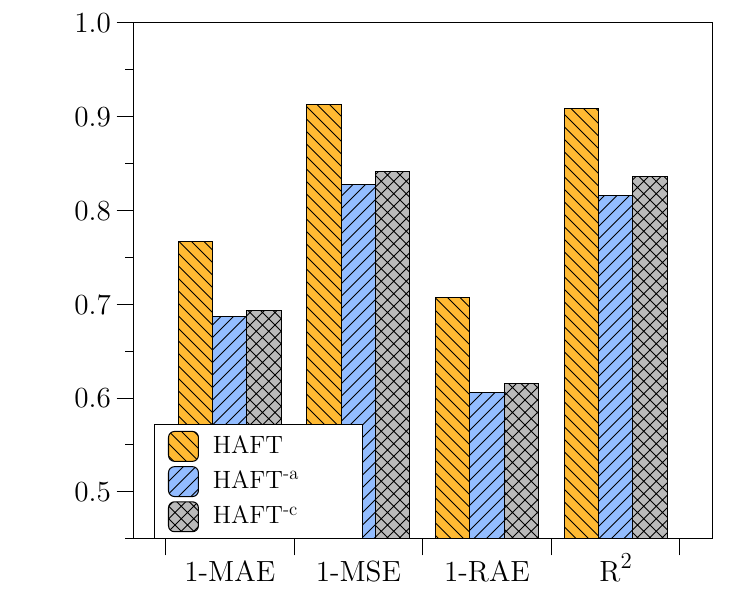}
        \caption{Openml\_586}
        \label{fig:Openml_586_critic_abla}
    \end{subfigure}
    \begin{subfigure}[t]{0.24\textwidth}
        \includegraphics[width=\textwidth]{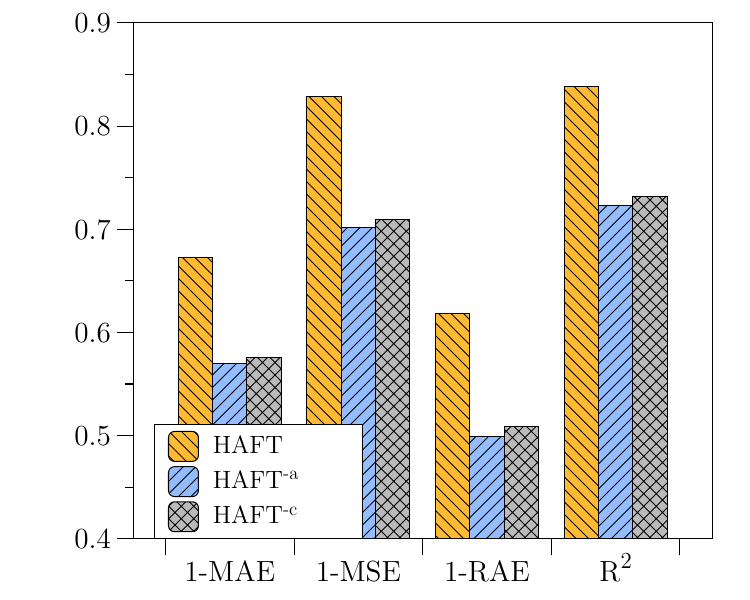}
        \caption{Openml\_620}
        \label{fig:Openml_620_critic_abla}
    \end{subfigure}
    \begin{subfigure}[t]{0.24\textwidth}
        \includegraphics[width=\textwidth]{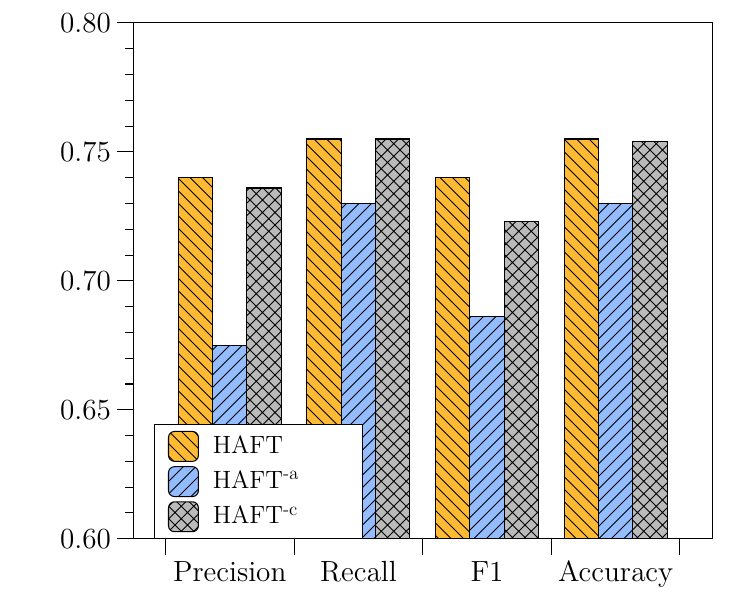}
        \caption{German Credit}
        \label{fig:German_credit_critic_abla}
    \end{subfigure}
    \begin{subfigure}[t]{0.24\textwidth}
        \includegraphics[width=\textwidth]{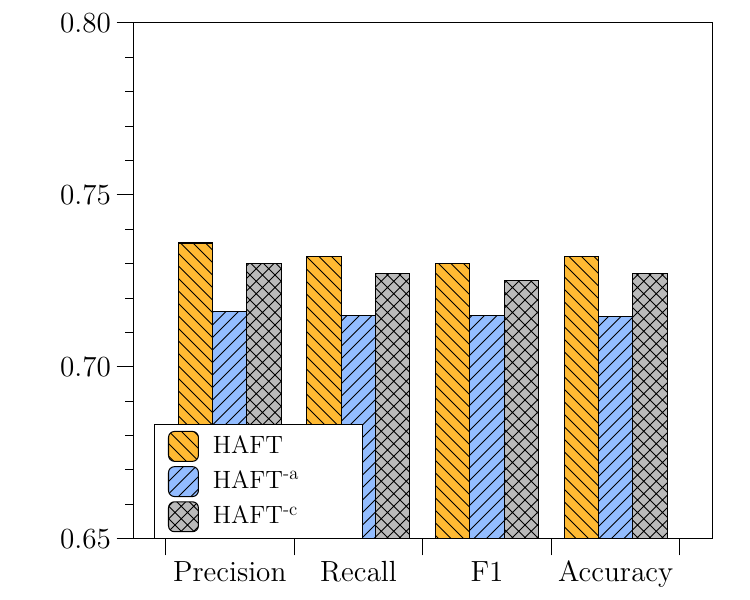}
        \caption{Messidor\_features}
        \label{fig:Messidor_feature_critic_abla}
    \end{subfigure}
    \vspace{-0.3cm}
  \caption{Influence of Shared Critic and Advantage Decomposition}
  \vspace{-0.4cm}
  \label{fig:critic_comparison}
  \Description{Effect of Shared Critic and Advantage Decomposition}
\end{figure*}

\section{Experiments}
 
\subsection{Experiment Setup}
\subsubsection{Datasets}
We use 23 publicly available datasets from UCI
\cite{uci_dataset}, LibSVM\cite{libsvm_dataset}, Kaggle\cite{kaggle_dataset}, and OpenML\cite{openml_dataset} to conduct our experiments. The 23 datasets involve 14 classification tasks and 9 regression tasks. Table~\ref{tab:overall_performance} reflects the statistics of the data.

\subsubsection{Evaluation Metrics}
We adopt standard evaluation metrics to ensure a comprehensive analysis of our model's performance.
Regression tasks were evaluated using the following metrics: 1-Relative Absolute Error (1-RAE) \cite{wang2022group}, 1-Mean Absolute Error (1-MAE), 1-Mean Squared Error (1-MSE) and coefficient of determination ($R^{2}$).
For classification tasks, assessment was conducted using Precision, Recall, F1-score and Accuracy. The formulae for the F1-score and 1-RAE are given by:
    $F_1 = 2 \cdot \frac{\text{Precision} \cdot \text{Recall}}{\text{Precision} + \text{Recall}}$ and 
$
    \text{1-RAE} = 1 - \frac{\sum_{i=1}^n |y_i - \tilde{y}_i|}{\sum_{i=1}^n |y_i - \bar{y}_i|},
$
where $y_i$, $\tilde{y}_i$, and $\bar{y}_i$ represent the ground truth, predictions, and the mean of the ground truth respectively.

\subsubsection{Baseline Model}
We compare our method with 8 widely-used feature transformation methods. 1) \textbf{RDG} generates feature-operation-feature transformation records randomly to create a new feature space. 
2) \textbf{ERG} applies operations to each feature to expand the feature space and selects the crucial features as the new feature set. 
3) \textbf{LDA} \cite{blei2003latent} is a matrix factorization-based method that obtains the factorized hidden state as the generated feature space. 
4) \textbf{AFAT} \cite{horn2020autofeat} is an enhanced version of ERG that repeatedly generates new features and uses multi-step feature selection to identify informative ones. 
5) \textbf{NFS} \cite{chen2019neural} models the transformation sequence of each feature and uses reinforcement learning (RL) to optimize the entire feature generation process. 
6) \textbf{TTG} \cite{khurana2018feature} formulates the transformation process as a graph and implements an RL-based search method to identify the best feature set. 
7) \textbf{GRFG} \cite{wang2022group} utilizes three collaborative reinforcement agents for feature generation and introduces a feature grouping strategy to accelerate agent learning. 
8) \textbf{DIFER} \cite{zhu2022difer} embeds randomly generated feature transformation records using a seq2seq model and employs gradient search to identify the best feature set.
Besides that, to validate the necessity of each component of \model, we develop three model variants: 
a. \textbf{\model$^{-c}$} replaces the shared critic in \model\ with separate critics for each agent;
b. \textbf{\model$^{-a}$} removes the advantage decomposition component from \model\ while retaining the shared critic.
c. \textbf{\model$^{-u}$} only utilize the  \(S_{stats}\) as the shared critic input. 
%We have made the code and data publicly available on GitHub, as mentioned in the Abstract section. 
%To further enhance reproducibility, we provide experimental platform and hyperparameter settings in Appendix~\ref{Environmental Settings and Reproducibility}.

\vspace{-0.2cm}
\subsubsection{Experimental Settings}
All experiments are conducted on an Ubuntu 22.04.5 LTS operating system, powered by an AMD Ryzen Threadripper PRO 7965WX 24-core processor, using Python 3.12.7 and PyTorch 2.5.1 as the framework.

\begin{figure*}[h]
  \centering
    \begin{subfigure}[t]{0.24\textwidth}
        \includegraphics[width=\textwidth]{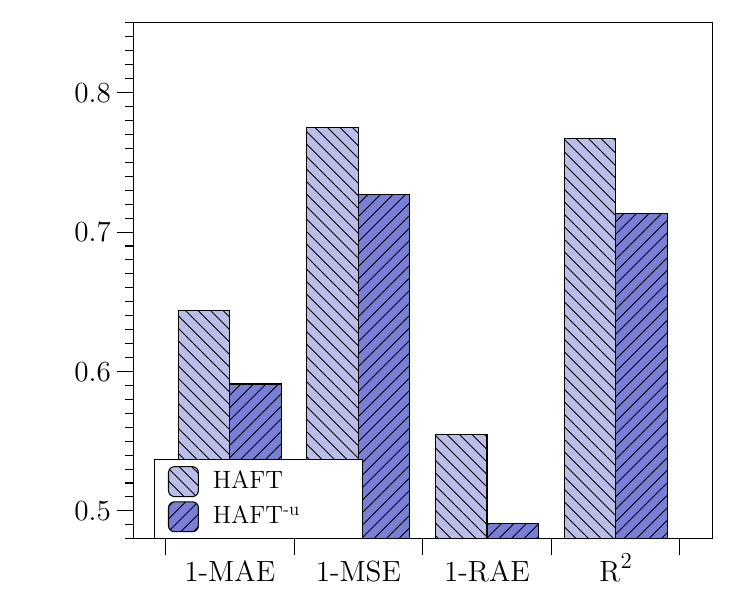}
        \caption{Openml\_589}
        \label{fig:Openml_589_critic_state}
    \end{subfigure}
    \begin{subfigure}[t]{0.24\textwidth}
        \includegraphics[width=\textwidth]{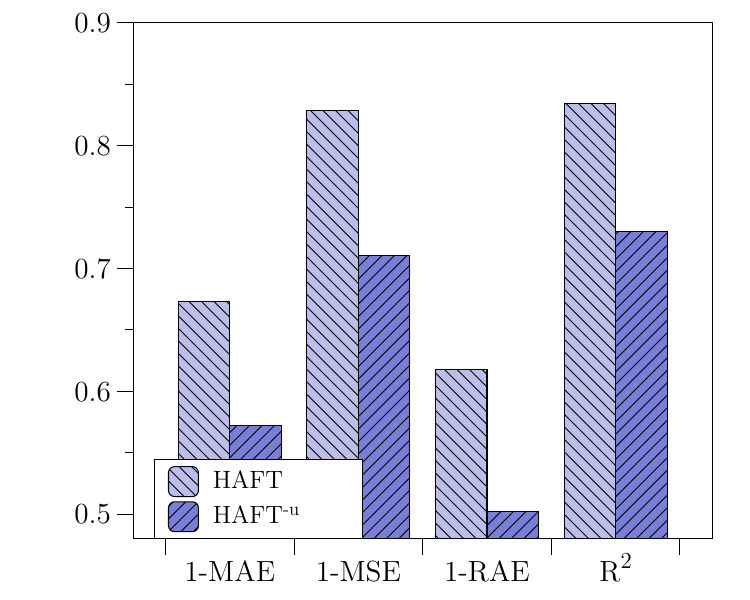}
        \caption{Openml\_620}
        \label{fig:Openml_620_critic_state}
    \end{subfigure}
    \begin{subfigure}[t]{0.24\textwidth}
        \includegraphics[width=\textwidth]{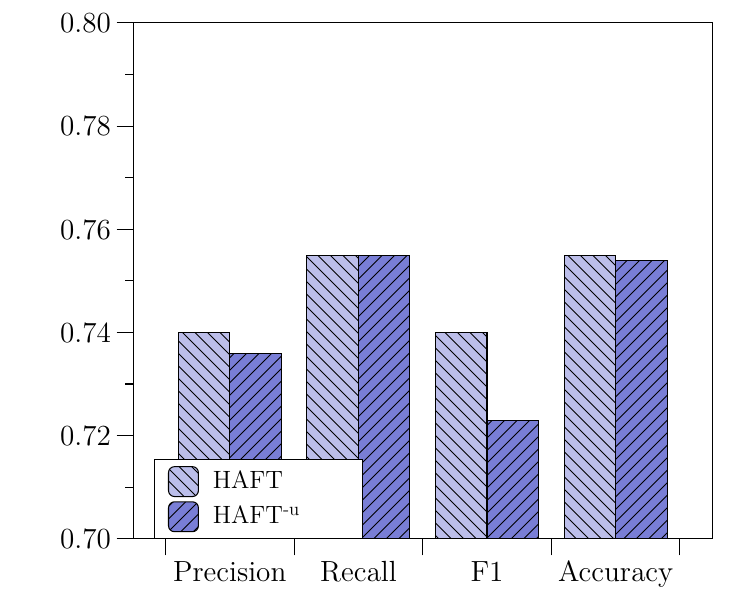}
        \caption{German Credit}
        \label{fig:German_credit_critic_state}
    \end{subfigure}
    \begin{subfigure}[t]{0.24\textwidth}
    \includegraphics[width=\textwidth]{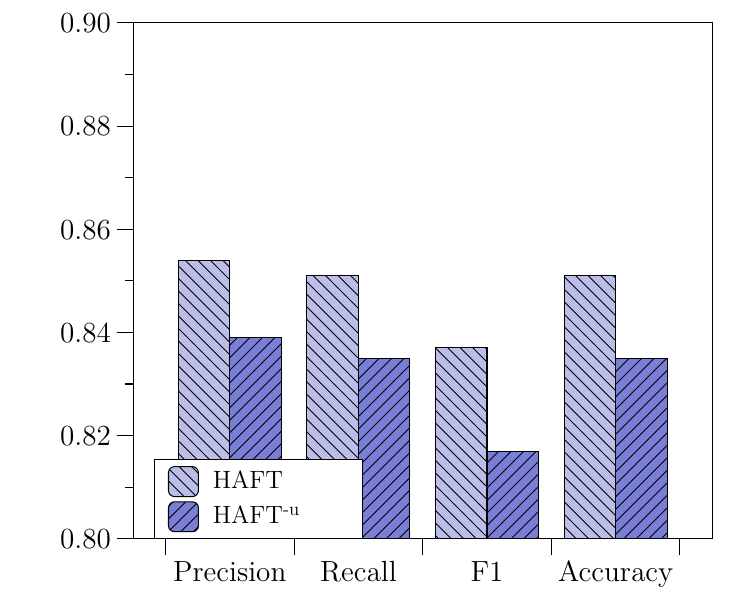}
        \caption{SVMGuide3}
        \label{fig:SVMguide3_critic_state}
    \end{subfigure}
    \vspace{-0.2cm}
  \caption{Influence of Different State Encoding}
  \vspace{-0.3cm}
  \label{fig:Influence of Different State Encoding}
  \Description{Comparison of value function training curves under different
  state encodings on four datasets: Openml\_589, Openml\_620, German credit,
  and SVMGuide3.}
\end{figure*}

\begin{figure*}[h]
\vspace{-0.1cm}
  \centering
    \begin{subfigure}[t]{0.24\textwidth}
        \includegraphics[width=\textwidth]{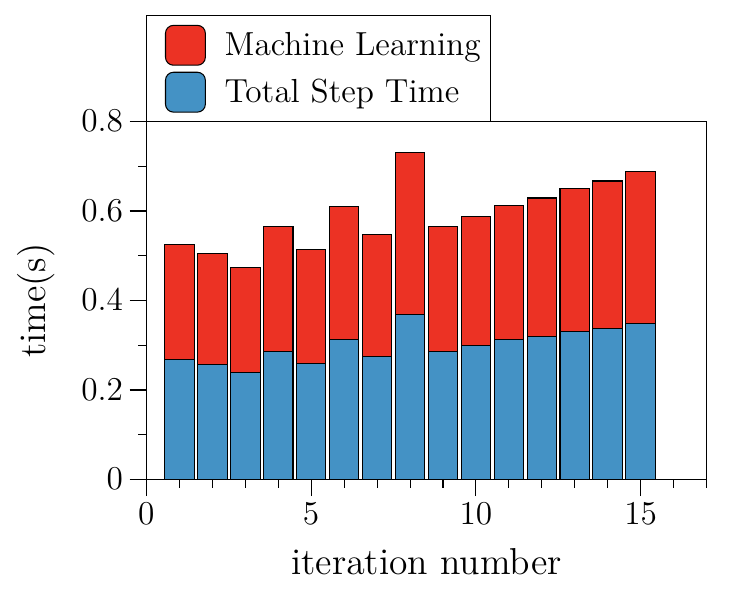}
        \caption{Housing Boston\_HAFT}
        \label{fig:housing_boston_tc}
    \end{subfigure}
    \begin{subfigure}[t]{0.24\textwidth}
        \includegraphics[width=\textwidth]{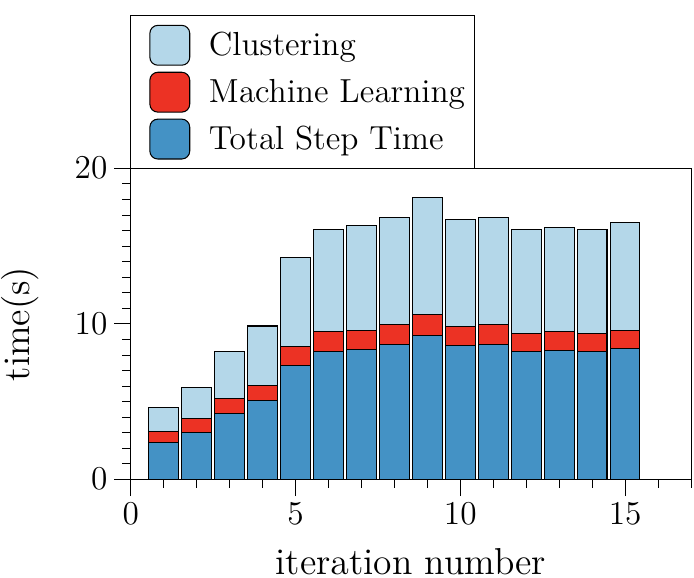}
        \caption{Housing Boston\_GRFG}
        \label{fig:housing_boston_GRFG_tc}
    \end{subfigure}
    \begin{subfigure}[t]{0.24\textwidth}
        \includegraphics[width=\textwidth]{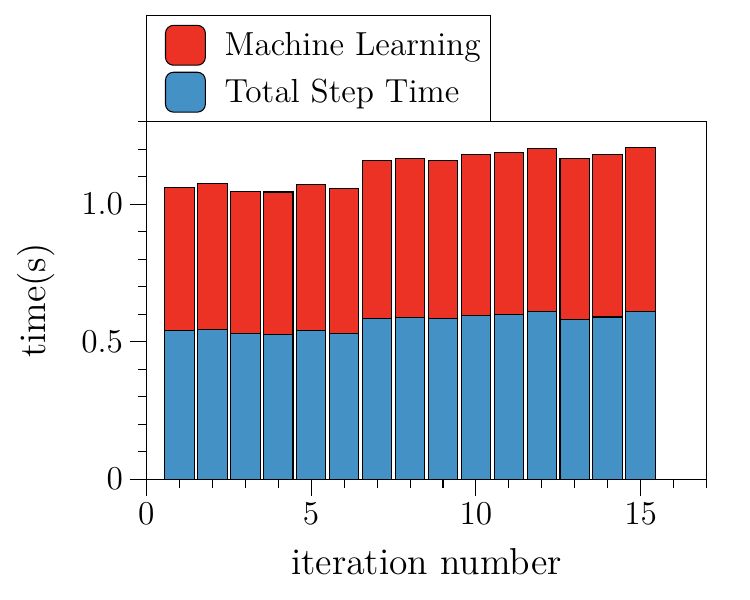}
        \caption{SpamBase\_HAFT}
        \label{fig:SpamBase_tc}
    \end{subfigure}
    \begin{subfigure}[t]{0.24\textwidth}
        \includegraphics[width=\textwidth]{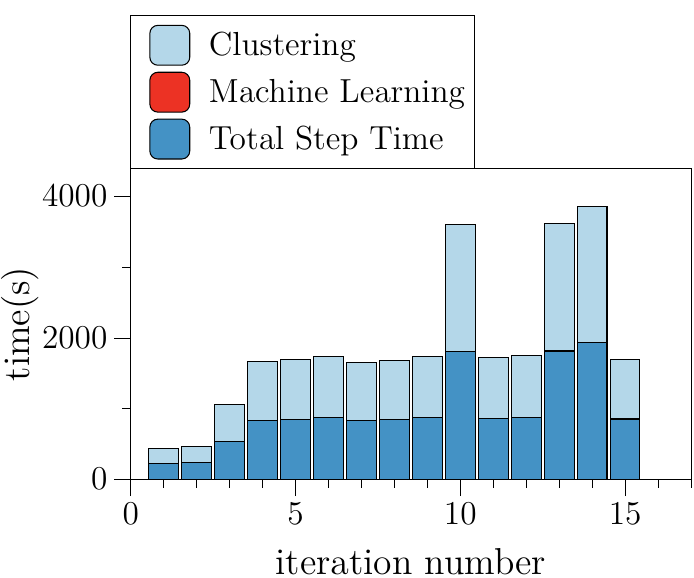}
        \caption{SpamBase\_GRFG}
        \label{fig:SpamBase_GRFG_tc}
    \end{subfigure}
\vspace{-0.2cm}
  \caption{Time Complexity Comparison between HAFT and GRFG}
  \vspace{-0.3cm}
  \label{Time_Complexity}
  \Description{Time Complexity between HAFT and GRFG}
\end{figure*}

\subsubsection{Hyperparameters and Reproducibility}
The operation set includes square root, square, cosine, sine, tangent, exponential, cube, logarithm, reciprocal, quantile transformation, min–max scaling, sigmoid, addition, subtraction, multiplication, and division. We cap the process at 100 episodes, each with up to 25 steps, and select the top 20 features using a mutual-information-based metric to evaluate downstream performance. The entropy coefficient ranges from 0.01 to 0.3, and we explore learning rates of 1e–3, 1e–4, and 1e–5 for both the feature agent and operation agent, the dimension for \(z_{\text{attn}}\)<=49 depending on the dataset.
In terms of network architecture, the feature agent employs two Transformer encoder layers and one MLP projection layer. Meanwhile, the shared critic consists of four MLP layers, while the operation agent comprises two MLP layers, each employing LeakyReLU as the activation function. The parameters of all baseline models are configured according to the default settings specified in the corresponding papers, ensuring consistency in evaluation settings \cite{ying2024unsupervised}.

\subsection{Experiment Results}

\subsubsection{Overall Performance}
This experiment aims to answer the following question: \textit{Can \model\ construct a better feature space to improve downstream task performance?} Table~\ref{tab:overall_performance} shows the experimental results in terms of F1 score or 1-RAE. (The bold/underline denotes best/second performance). 
We observe that \model\ consistently outperforms other baselines across most domains and tasks while maintaining a stable top ranking.
There are two potential reasons for the experimental observation:
1) the multi-head attention-based feature agent structure enables feature agents to focus on more significant features, leading to the generation of new informative features.
2) the shared critic mechanism improves communication and coordination among agents, facilitating the learning of a more effective feature transformation policy.
Another notable observation is that, unlike \model, the other models fail to maintain a stable performance rank throughout our experiments.
This observation reflects the effectiveness of \model\ in learning superior transformation strategies compared to other methods.
In summary, this experiment demonstrates that \model\ effectively refines the feature space to further enhance downstream tasks.

\vspace{-0.1cm}
\subsubsection{Ablation Study 1: Shared Critic}
This experiment aims to answer: \textit{Does a shared critic and advantage decomposition update improve agent communication and lead to more effective feature transformation policies over separate critics?}
To answer this question, we develop model variant \model$^{-c}$, which replaces the shared critic with separate critics for each agent, and \model$^{-a}$ which removes the advantage decomposition component from
\model\ while retaining the shared critic.
We report the comparison results in terms of F1 score or 1-RAE on two regression datasets  (e.g., Openml\_586, Openml\_620) and two classification datasets (e.g., German Credit, Messidor\_features). Figure~\ref{fig:critic_comparison} presents the experimental results in terms of corresponding evaluation metrics.
We find that \model\ consistently beats \model$^{-c}$ and \model$^{-a}$ across various datasets, and \model$^{-a}$ and \model$^{-c}$ perform on par with each other. 
This observation can be explained by two key mechanisms: The shared critic provides global guidance, and there might also exist updating direction conflict, while the advantage decomposition scheme enhances cooperation through more informative reward signals.
This leads to the development of reliable feature-transformation policies, resulting in superior feature spaces. In summary, this experiment shows that the shared critic is essential for performance.
%improving feature transformation.

\begin{figure*}[h]
\vspace{-0.1cm}
  \centering
    \begin{subfigure}[t]{0.24\textwidth}
        \includegraphics[width=\textwidth]{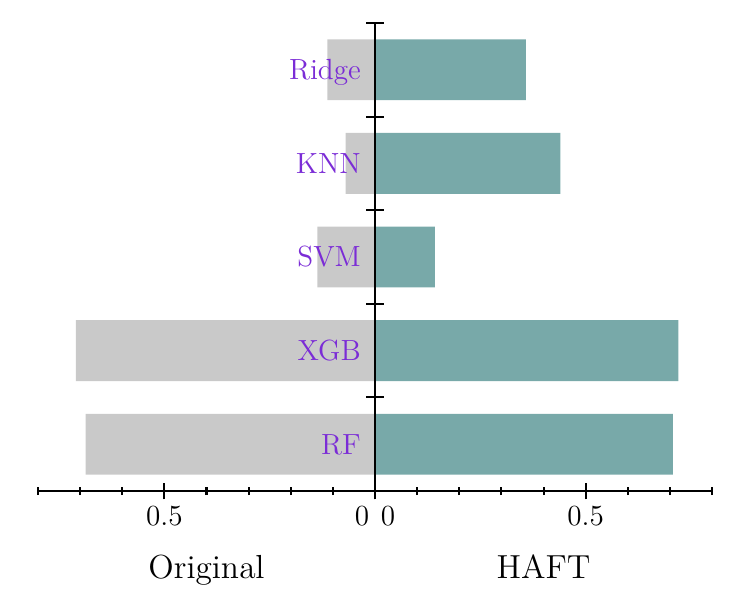}
        \caption{Openml\_586}
        \label{fig:Openml_586_rb}
    \end{subfigure}
    \begin{subfigure}[t]{0.24\textwidth}
        \includegraphics[width=\textwidth]{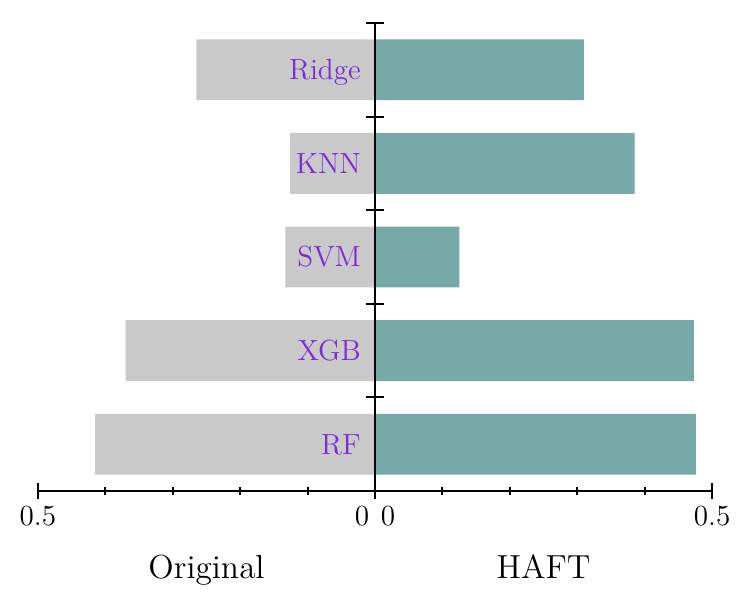}
        \caption{Openml\_618}
        \label{fig:Openml_618}
    \end{subfigure}
    \begin{subfigure}[t]{0.24\textwidth}
        \includegraphics[width=\textwidth]{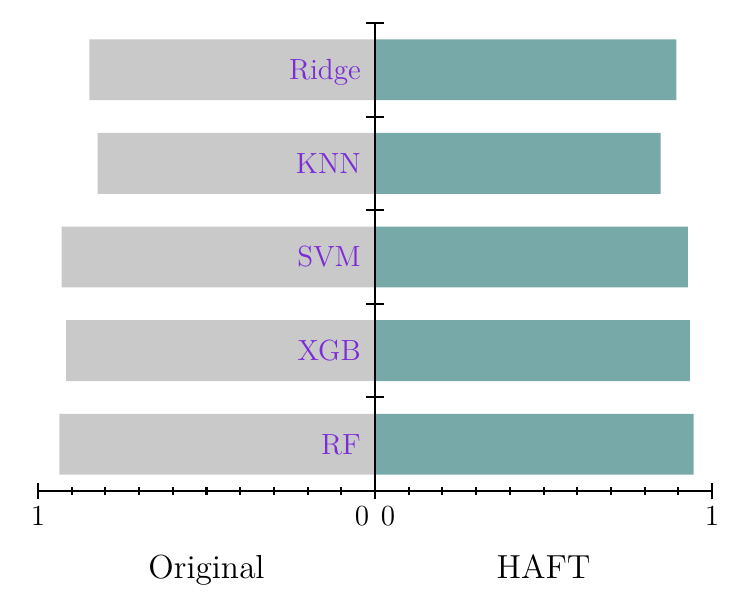}
        \caption{Ionosphere}
        \label{fig:Ionosphere}
    \end{subfigure}
    \begin{subfigure}[t]{0.24\textwidth}
        \includegraphics[width=\textwidth]{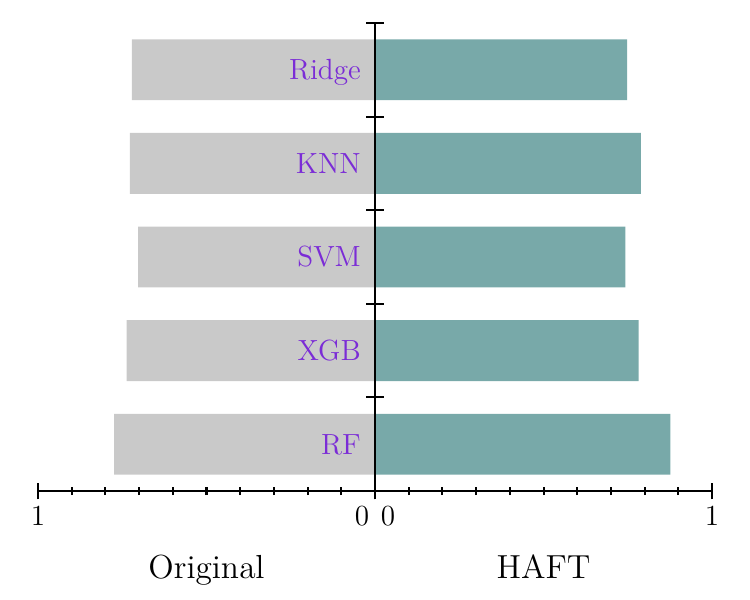}
        \caption{SpectF}
        \label{fig:SpectF_rb}
    \end{subfigure}
    \vspace{-0.2cm}
  \vspace{-0.1cm}
  \caption{Robustness Check of \model\ when confronted with different downstream ML models in terms of F1 score and 1-RAE }
  \vspace{-0.3cm}
  \label{fig:robustness}
  \Description{An overview of our MARL-based feature selection Framework}
\end{figure*}

\begin{figure}[ht]
  \centering
    \begin{subfigure}[t]{0.49\linewidth}
        \includegraphics[width=\textwidth]{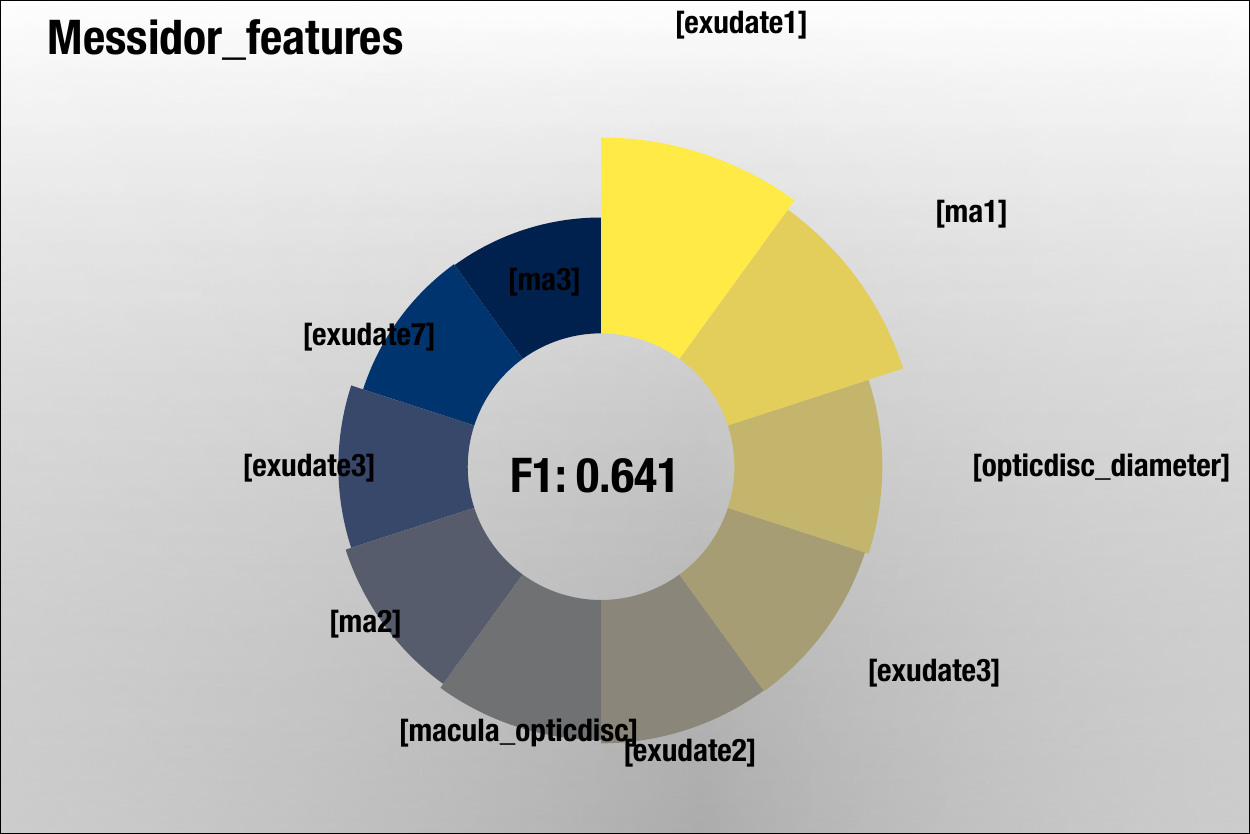}
        \caption{Original Feature Space}
        \label{fig:messidor_case_original}
    \end{subfigure}
    \begin{subfigure}[t]{0.49\linewidth}
        \includegraphics[width=\textwidth]{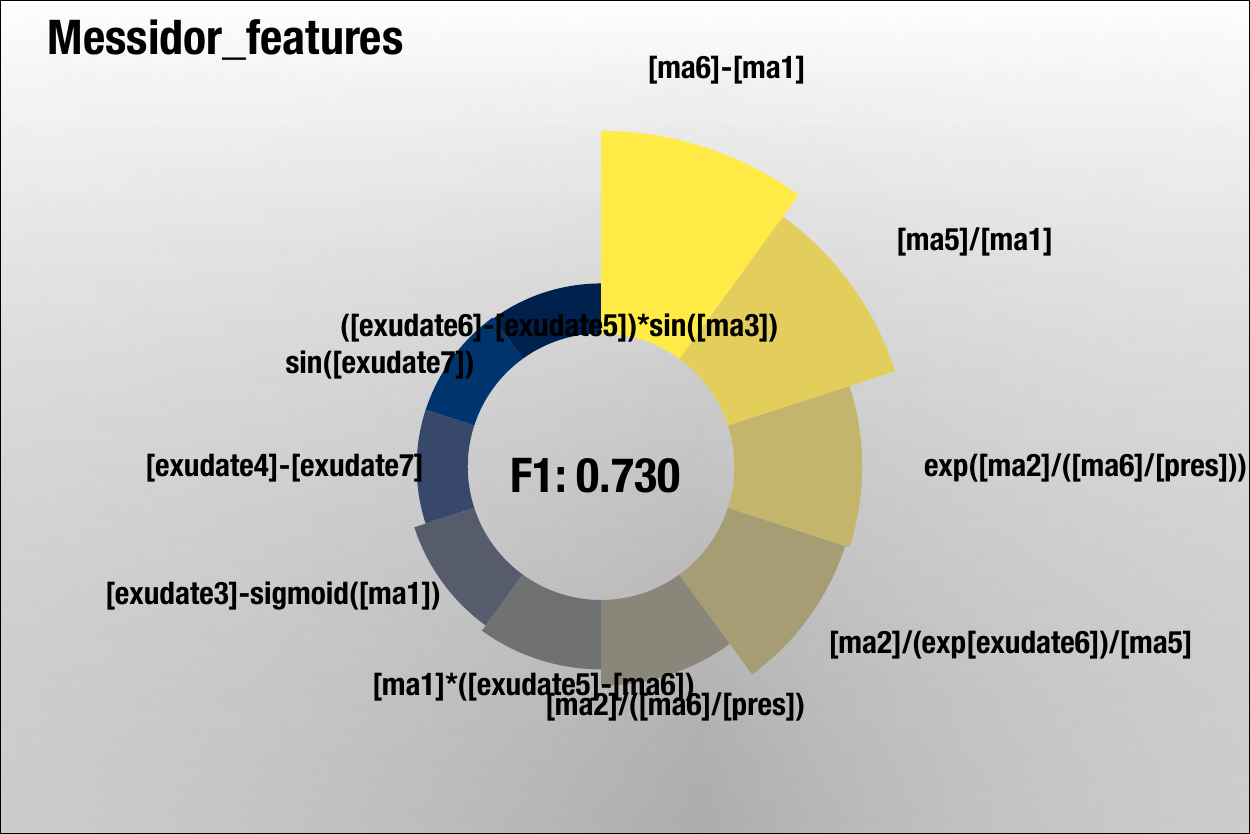}
        \caption{Reconstructed Feature Space}
        \label{fig:messidor_case_HAFT}
    \end{subfigure}
    \vspace{-0.2cm}
  \vspace{-0.1cm}
  \caption{Case Study of Messidor\_features dataset}
  \label{fig:Case_Study}
  \Description{An overview of our MARL-based feature selection Framework}
  \vspace{-0.1cm}
\end{figure}
%\vspace{-0.5cm}

\subsubsection{Ablation Study 2: 
Shared Critic States }
This experiment aims to answer: \textit{Does incorporating interaction information into the shared critic state improve feature transformation performance?}
To address this question, we create the model variant \model$^{-u}$ by removing the interaction component of our shared critic's state input.
We then compare the performance of \model\ and \model$^{-u}$ across two regression tasks (i.e., Openml\_589 and Openml\_620) and two classification tasks (i.e., German Credit and SVMGuide3).
Figure~\ref{fig:Influence of Different State Encoding} shows the comparison results in terms of F1 score or 1-RAE. 
We observe that \model\ surpasses \model$^{-u}$ in most cases across all metrics. 
A plausible reason is that appending interaction information, alongside statistical representations of the feature space, helps to enrich the input to the shared critic, enabling it to better assess the quality of transformed features and guide the agent system more effectively.
In summary, this experiment shows that incorporating interaction information is a beneficial design choice in \model.

\subsubsection{Robustness Check}
This experiment aims to answer: \textit{Does \model\ exhibit robustness when confronted with various machine learning models serving as downstream tasks?} 
To answer this question, we replace the downstream ML model with Random Forest (RF)~\cite{breiman2001random}, XGBoost (XGB)~\cite{chen2016xgboost}, Support Vector Machine (SVM)~\cite{cortes1995support}, K-Nearest Neighbors (KNN)~\cite{cover1967nearest}, and Ridge Regression (Ridge)~\cite{hoerl1970ridge}, respectively. Figure~\ref{fig:robustness} presents the comparison results in terms of F1 score or 1-RAE on Openml\_586, Openml\_618, Ionosphere, and SpectF datasets. We find that regardless of changes in downstream models, \model\ consistently refines the feature space and achieves better results than the original feature space.
The underlying reason is that \model\ can customize feature transformation policies to the specific characteristics of downstream models, resulting in more effective feature spaces.
Thus,  this experiment demonstrates the robustness of \model\ across different downstream models.
%in producing effective feature spaces across different downstream machine learning models.

\subsubsection{Scalability Analysis}

This experiment aims to address the question: \textit{Does \model\ exhibit better scalability than GRFG in refining the feature space?}
GRFG is the algorithm most similar to \model, utilizing a group-wise feature generation strategy to refine the feature space. 
To further analyze scalability, we conduct experiments to compare the time costs of key components in GRFG and \model\ over 15 iterations on the Housing Boston and SpamBase datasets. 
Figure~\ref{Time_Complexity} presents the results of the time complexity comparison.
We observe that the time cost per iteration of \model\ is significantly lower than that of GRFG, with GRFG requiring significantly more time. 
The primary reason for this observation is that clustering is the most time-consuming step in GRFG, as shown in Figure~\ref{Time_Complexity}.
Even with the high time cost in GRFG, its performance is still worse than \model. 
This observation suggests that although GRFG's clustering strategy aims to refine the feature space group-wise, many of the generated features are ineffective for refinement. 
In contrast, \model\ utilizes multi-head attention to capture complicated feature interactions to select suitable features for feature crossing.
Thus, \model\ is much more scalable and applicable in realistic scenarios.

\begin{comment}

\end{comment}

\subsubsection{Case Study}
This experiment aims to answer:\textit {Can \model\ generate an explainable and traceable feature space?} 
We select the top 10 essential features using mutual information as rank criteria for prediction in both the original and \model\ transformed feature spaces of the Messidor\_features dataset for comparison. Figure~\ref{fig:Case_Study} shows the comparison results, where larger pie slices indicate higher feature importance. We find that all of the critical features in the new feature space were generated by \model\, resulting in a 13.9\% improvement in downstream machine learning performance. This indicates that \model\ transforms features to refine the feature space. We also observe that one of the key features in the newly generated feature space is derived from the original feature ['ma1']. This indicates that \model\ is capable of tracing and explaining the transformed feature by checking its name. In our case, ['ma1'] indicates MA detection at a 0.5 confidence level, and ['\verb|[ma6]-[ma1]|'] and ['\verb|[ma5]/[ma1]|'] reveals a traceable, non-linear relationship with diabetic retinopathy diagnosis.
Thus, the experiment shows the effectiveness of \model\ from another angle.

\vspace{-0.1cm}
\subsubsection{Short‑ vs. Long‑Term Gain Trade‑Off}
% ----------------------------------------
This experiment aims to answer: \textit{How does the trade-off between short-term and long-term gains affect the sequential decision-making process of \model?}
We select the Airfoil dataset as a candidate for this case study and demonstrate a 5-step feature transformation process. 
Table \ref{tab:feat_steps} shows the exploration process with each step's immediate and cumulative performance gains.
We observe that in step 2, the transformation of $U_{\infty}$ into $\sigma\!\bigl(U_{\infty}\bigr)$ causes the immediate gain to decrease by $-0.2\%$, while in step 4, this transformation contributes to improving the cumulative gain by a substantial $2.5\%$. 
A plausible explanation is that our multi-agent reinforcement learning framework expands the exploration depth during the search process, while the reward-guided schema directs the search to focus on cumulative gains rather than short-sighted decisions.
This experiment demonstrates that our collaborative multi-agent system enables the model to balance short-term and long-term gains, allowing sophisticated feature space exploration where initial transformation steps may temporarily reduce performance but yield better overall results.

\subsubsection{Study of the Hyperparameter Sensitivity of \model\ }
\label{hyperparameter sensitivity}
\begin{figure}[h]
  \centering
    \begin{subfigure}[t]{0.49\linewidth}
        \includegraphics[width=\linewidth]{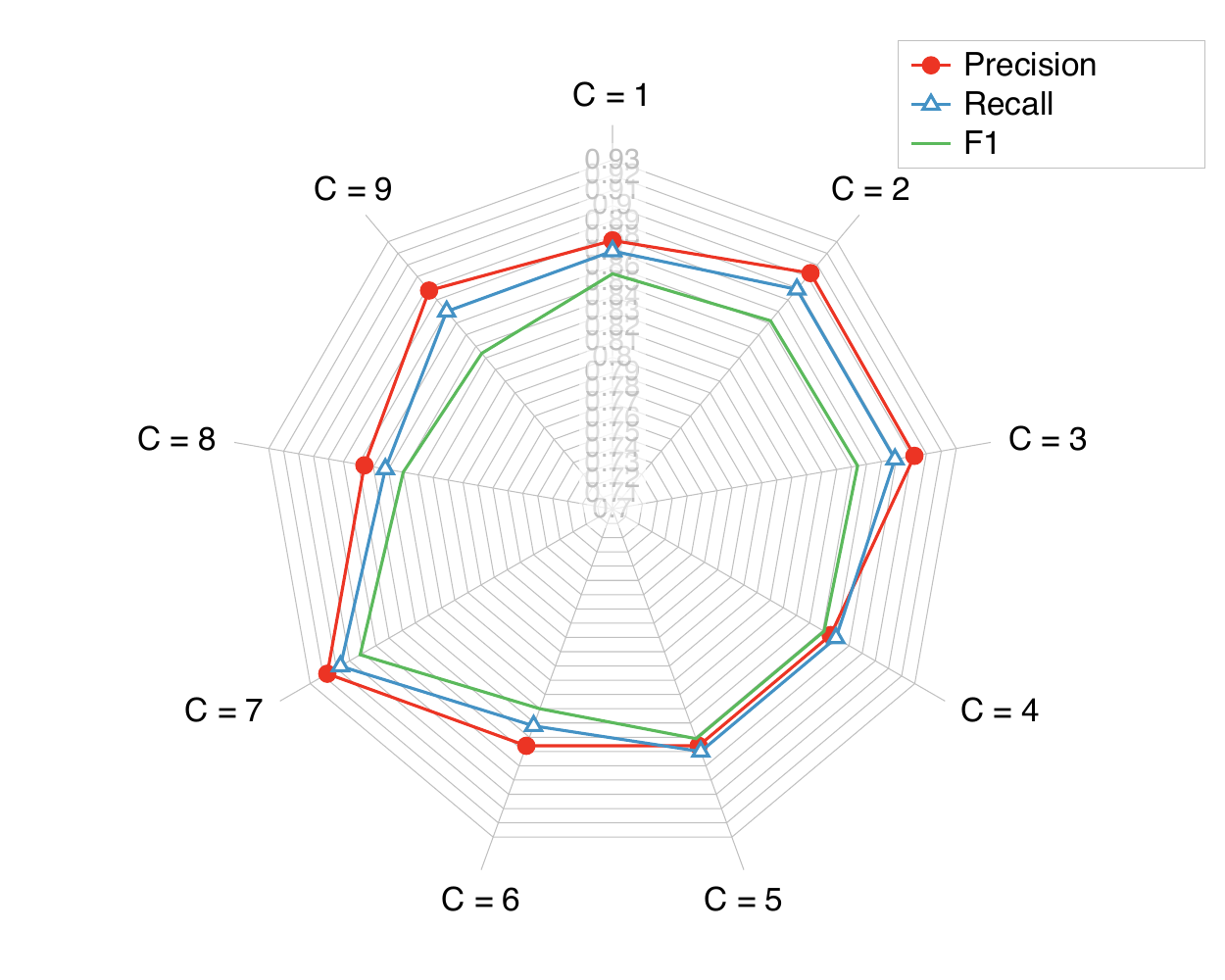}
        \caption{Number of Transformer Layers $C$ (under $M$ = 15)}
        \label{fig:Openml_586_param}
    \end{subfigure}
    \begin{subfigure}[t]{0.49\linewidth}
        \includegraphics[width=\linewidth]{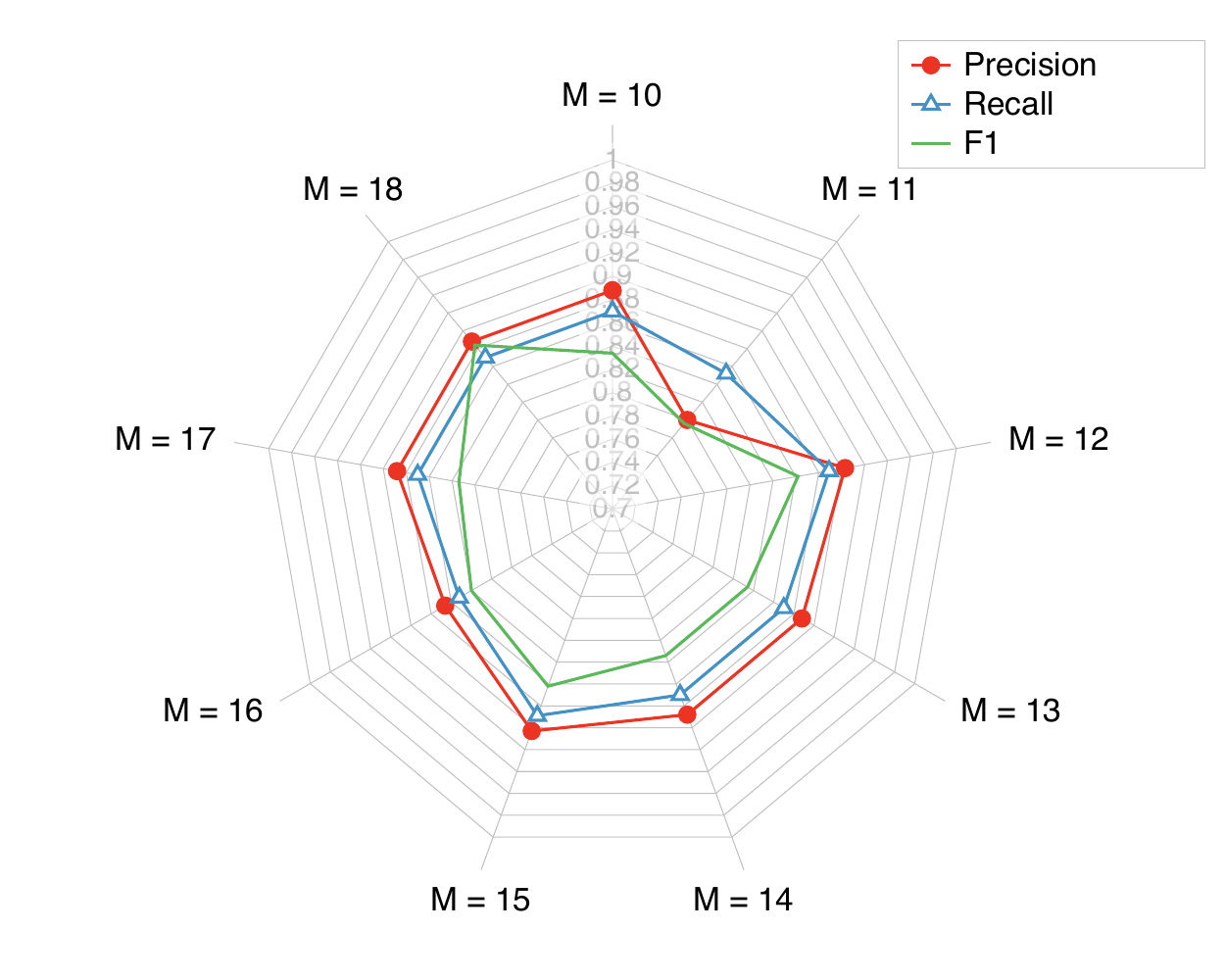}
        \caption{Steps per Episode $M$ (under $C$ = 2) }
        \label{fig:Openml_589_params}
    \end{subfigure}
  \vspace{-0.3cm}
  \caption{The hyperparameter sensitivity test on SpectF}
  \label{fig:Hyperparamter}
  \Description{: Parameter sensitivity}
\end{figure}

\noindent This experiment aims to answer: \textit{To what extent is the performance of \model\ sensitive to the transformer encoder layer number $C$ and the step size $M$?} 
We configured the transformer encoder layer number $C$ from 1 to 9 and the step size $M$ from 10 to 18, then trained \model\ on the SpectF dataset. The performance metrics, including Precision, Recall, and F1, are reported in Figure~\ref{fig:Hyperparamter}.
We observe that \model\ is not sensitive to encoder layer numbers. The plausible explanation is that a single layer of the transformer encoder is sufficiently complex to capture interactions between features and make informed decisions. Additionally, we find that \model\ is robust to training step size. The underlying reason for this observation is that \model\ can transform informative features without simplifying or complicating the transformation due to variations in the number of steps. And it tends to make decisions based on feature quality rather than the number of operations allowed.
Thus, this experiment demonstrates that \model\ is not sensitive to these distinct parameter settings, the learning process of \model\ is relatively robust and stable.
\vspace{-0.3cm}

\begin{table}[htbp]
  \centering
  \caption{ Short- vs. Long-Term Gain Trade-Off}
  \vspace{-0.3cm}
  \label{tab:feat_steps}
  \begin{tabular}{@{}c l l l l r r@{}}
    \toprule
    Step & $f_{1}$ & Op & $f_{2}$ & New
         & {$\Delta$(\%)\textsuperscript{a}}
         & {Cum.\,(\%)}\textsuperscript{b} \\
    \midrule
     1 & $c$           & $\sin$    & -- & $\sin(c)$                           & +0.9\% & +0.9\% \\
     2 & $U_{\infty}$  & $\sigma$  & -- & $\textcolor{green}{\sigma\!\bigl(U_{\infty}\bigr)}$    & \textcolor{blue}{–0.2\%} & +0.7\% \\
     3 & $\delta$      & $\exp$    & -- & $\exp(\delta)$                      & +1.8\% & +2.5\% \\
     4 & $\sin(c)$     & /         & $\sigma(U_{\infty})$
                                      & $\dfrac{\sin(c)}{\textcolor{green}{\sigma(U_{\infty})}}$ & \textcolor{red}{+2.5\%} & +5.0\% \\
     5 & $\alpha$      & square    & -- & $\alpha^{2}$                        & +1.3\% & +6.3\% \\
     \midrule
     \multicolumn{7}{c}{\dots} \\
    \bottomrule
  \end{tabular}
  % ------------ notes, now left‑aligned -------------
  \begin{flushleft}
    \footnotesize
    \textsuperscript{a}\,Immediate change in validation accuracy.\par
    \textsuperscript{b}\,Cumulative change from the baseline.
  \end{flushleft}
\end{table}
\vspace{-0.6cm}

\vspace{-0.3cm}

% \vspace{-0.2cm}
\section{Related Works}

\textbf{Automated Feature Transformation}
aims to refine or augment the original feature space so that machine learning models can more effectively capture complex, high-order relationships among variables. Traditionally, feature transformation techniques fall into two broad categories: latent representation learning-based and feature transformation-based approaches. Latent representation learning-based methods, as demonstrated in ~\cite{bengio2013representation,guo2017deepfm}, focus on learning an optimized, often high-dimensional latent space where feature interactions are implicitly encoded. While these methods are known to boost predictive accuracy, the resulting representations can be challenging to interpret or directly trace back to the original input space. 
In contrast, feature transformation-based methods~\cite{wang2022group,huang2024enhancing,hu2024reinforcement,xiao2024traceable,azim2024feature} generate new features using arithmetic or aggregate operations. Although these methods yield traceable features, they often encounter an exponentially growing search space problem and become inefficient when dealing with large feature sets. In contrast to previous approaches, our attention-based feature crossing strategy not only captures the inherent relationships among features but also scales efficiently. This is achieved by employing cascading agents guided by a shared critic to learn effective feature interaction policies.

\noindent \textbf{Reinforcement Learning} investigates how intelligent agents should act within an environment to maximize the expected cumulative reward~\cite{sutton2018reinforcement,wiering2012reinforcement,wang2026optimizing}. Based on the learned policy, reinforcement learning algorithms can be classified into two main categories: value-based and policy-based. Value-based algorithms~\cite{mnih2013playing,van2016deep} estimate the value of states or state-action pairs to facilitate action selection, whereas policy-based algorithms~\cite{sutton1999policy} learn a probability distribution that maps states to actions. Furthermore, the actor-critic framework~\cite{schulman2017proximal} combines the strengths of both approaches.
While single-agent reinforcement learning has seen remarkable success, recent work increasingly focuses on multi-agent settings~\cite{wen2022multi} where multiple agents collaborate toward a shared goal. However, variations in observations, actions, or rewards often arise, making effective communication critical for coordination and overall performance.
Foerster et al.~\cite{foerster2016learning} introduced RIAL and DIAL to facilitate communication among agents. 
Building on this, Foerster et al.~\cite{foerster2018counterfactual} proposed the COMA, employing a centralized critic and a counterfactual baseline for credit assignment. This design supports effective coordination even when agents have distinct reward functions.
Lowe et al.~\cite{lowe2017multi} extended the actor-critic paradigm by implementing a centralized critic for each agent and allowing them to learn continuous policies, improving the scalability and flexibility of MARL systems.
Moreover, Lyu et al.~\cite{lyu2021contrasting} found that centralized critics do not consistently outperform their decentralized counterparts across various evaluation domains.
Recent advances in the MARL field, such as the theoretical formulation and empirical adaptation of PPO to cooperative multi-agent settings~\cite{yu2022surprising,kuba2021trust}, and Asynchronous Action Coordination~\cite{zhang2024sequential}, have accelerated research progress in this area.
Overall, these studies underscore the importance of communication, collaboration, and effective credit assignment in the context of MARL. In our work, we adopt a similar communication mechanism by introducing a customized cascading design alongside a shared critic to address our feature transformation task.

%\vspace{-0.4cm}
\section{Conclusion}
In this paper, we propose a heterogeneous multi-agent framework for automated feature transformation, designed to effectively refine the feature space and enhance downstream task performance.
The core of this framework is a cascading agent structure consisting of two feature agents and one operation agent, which collaborate to explore feature transformation strategies and optimize the feature space.
We customize heterogeneous agent structures for selecting candidate features and operations, enabling the agents to learn more intelligent and reliable policies for feature transformation.
To further enhance communication among the agents, we develop a shared critic structure that evaluates each agent's decisions based on global feature space information and the decisions of other agents.
To adapt to the dynamically expanding feature space, we implement a multi-head attention-based agent structure to efficiently select suitable features.
Besides that, we propose a state encoding technique to enhance the learning process of RL agents.
Extensive experiments demonstrate that \model\ effectively refines the original feature space while exhibiting excellent scalability. The shared critic mechanism plays a crucial role in enhancing agent communication, leading to improved policies. Moreover, state encoding enhances the performance of reinforcement learning, enabling the generation of effective feature crossing strategies. 
In the future, enhancing generalization and adapting to dynamic data environments will be promising directions for further exploration.

% \begin{acks}
% \end{acks}

\bibliographystyle{ACM-Reference-Format}
\bibliography{sample-base,Tao}

\newpage

\end{document}